\documentclass[10pt,letterpaper]{article}

\usepackage[top=0.85in,left=3.3cm,right=3.65cm,footskip=0.75in]{geometry}

\usepackage{changepage}

\usepackage[utf8]{inputenc}

\usepackage{textcomp,marvosym}

\usepackage{fixltx2e}

\usepackage{amsmath,amssymb}

\usepackage{graphicx}
\usepackage[export]{adjustbox}

\usepackage{cite}

\usepackage{nameref,hyperref}

\usepackage{microtype}
\DisableLigatures[f]{encoding = *, family = * }

\usepackage{rotating}

\raggedright
\usepackage[aboveskip=1pt,labelfont=bf,labelsep=period,justification=raggedright,singlelinecheck=off]{caption}

\makeatletter
\renewcommand{\@biblabel}[1]{\quad#1.}
\makeatother

\date{}

\usepackage{siunitx}
\usepackage{enumitem}

\usepackage{bbm}

\usepackage{acronym} 
\acrodef{LIF}{Leaky Integrate-and-Fire}
\acrodef{STDP}{Spike-Timing-Dependent Plasticity}
\acrodef{PSP}{Postsynaptic Potential}
\acrodef{SRM}[SRM\textsubscript{0}]{Spike Response Model}
\acrodef{vRD}{van Rossum Distance}
\acrodef{VPD}{Victor \& Purpura Distance}
\acrodef{ReSuMe}{Remote Supervised learning Method}
\acrodef{SNN}{Spiking Neural Network}
\acrodef{PSD}{Precise-Spike-Driven}
\acrodef{INST}{INSTantaneous error}
\acrodef{SPAN}{Spike Pattern Association Neuron}
\acrodef{FILT}{FILTered error}
\acrodef{CHRON}{Chronotron}
\acrodef{FP}{Finite Precision}
\acrodef{HTP}{High-Threshold Projection}
\acrodef{MPDP}{Membrane Potential Dependent Plasticity}

\begin{document}
\vspace*{0.35in}

\begin{flushleft}
{\bf \Large
\textbf\newline{Supervised Learning in Spiking Neural Networks for Precise Temporal Encoding}
}
\newline
\\
{\bf Brian Gardner\textsuperscript{*},
André Grüning}
\\
Department of Computer Science, 
University of Surrey, Guildford, GU2 7XH, U.K.
\\
* E-mail: b.gardner@surrey.ac.uk
\end{flushleft}

\section*{Abstract}
Precise spike timing as a means to encode information in neural networks is biologically supported, and is advantageous over frequency-based codes by processing input features on a much shorter time-scale. For these reasons, much recent attention has been focused on the development of supervised learning rules for spiking neural networks that utilise a temporal coding scheme. However, despite significant progress in this area, there still lack rules that have a theoretical basis, and yet can be considered biologically relevant. Here we examine the general conditions under which synaptic plasticity most effectively takes place to support the supervised learning of a precise temporal code. As part of our analysis we examine two spike-based learning methods: one of which relies on an instantaneous error signal to modify synaptic weights in a network (INST rule), and the other one relying on a filtered error signal for smoother synaptic weight modifications (FILT rule). We test the accuracy of the solutions provided by each rule with respect to their temporal encoding precision, and then measure the maximum number of input patterns they can learn to memorise using the precise timings of individual spikes as an indication of their storage capacity. Our results demonstrate the high performance of the FILT rule in most cases, underpinned by the rule's error-filtering mechanism, which is predicted to provide smooth convergence towards a desired solution during learning. We also find the FILT rule to be most efficient at performing input pattern memorisations, and most noticeably when patterns are identified using spikes with sub-millisecond temporal precision. In comparison with existing work, we determine the performance of the FILT rule to be consistent with that of the highly efficient E-learning Chronotron rule, but with the distinct advantage that our FILT rule is also implementable as an online method for increased biological realism.

\section*{Introduction}
It is becoming increasingly clear that the relative timings of spikes transmitted by neurons, and not just their firing rates, is used to convey information regarding the features of input stimuli \cite{VanRullen2005}. Spike-timing as an encoding mechanism is advantageous over rate-based codes in the sense that it is capable of tracking rapidly changing features, for example briefly presented images projected onto the retina \cite{Gollisch2008} or tactile events signalled by the fingertip during object manipulations \cite{Johansson2004}. It is also apparent that spikes are generated with high temporal precision, typically on the order of a few milliseconds under variable conditions \cite{Mainen1995,Reich1997,Uzzell2004}.

The indicated importance of precise spiking as a means to process information has motivated a number of theoretical studies on learning methods for \acp{SNN} (reviewed in \cite{Kasinski2006,Gutig2014}). Despite this, there still lack supervised learning methods that can combine high technical efficiency with biological plausibility, as well as those claiming a solid theoretical foundation. For example, while the previously proposed \acf{SPAN} \cite{Mohemmed2012} and \acf{PSD} \cite{Yu2013} rules have both demonstrated success in training \acp{SNN} to form precise temporal representations of spatio-temporal spike patterns, they have lacked analytical rigour during their formulation; like many existing supervised learning methods for \acp{SNN}, these rules have been derived from a heuristic, spike-based reinterpretation of the Widrow-Hoff learning rule, therefore making it difficult to predict the validity of their solutions in general.

The E-learning \acl{CHRON} \cite{Florian2012} has emerged as a supervised learning method with stronger theoretical justification, considering that it instead works to minimise an error function based on the \ac{VPD} \cite{Victor1996}; the \ac{VPD} is a metric for measuring the temporal difference between two neural spike trains, and is determined by computing the minimum cost required to transform one spike train into another via the addition, removal or temporal-shifting of individual spikes. In this study, two supervised learning rules were formulated: the first termed E-learning, which is specifically geared towards classifying spike patterns using precisely-timed output spikes, and which provides high network capacity in terms of the number of memorised patterns. The second rule is termed I-learning, which is more biologically plausible than E-learning but comes at the cost of a reduced network memory capacity. The E-learning rule has less biological relevance than I-learning given its restriction to offline-based learning, as well as its dependence on synaptic variables that are non-local in time. Further, analytical, spike-based learning methods have been proposed in \cite{Memmesheimer2014}, such as the \acl{HTP} rule, and have demonstrated very high network capacity, but these similarly have been restricted in their implementation to offline learning.

A probabilistic method which optimises by gradient ascent the likelihood of generating a desired output spike train has been introduced by Pfister et al. in \cite{Pfister2006}. This supervised method has strong theoretical justification, and importantly has been shown to give rise to synaptic weight modifications that mimic the results of experimental \ac{STDP} protocols measuring the change in synaptic strength, triggered by the relative timing differences of pre- and postsynaptic spikes \cite{Bi1998}. Furthermore, the statistical framework in which this method has been devised is general, allowing for its extension to diverse learning paradigms such as reinforcement-based learning \cite{Fremaux2010}, backpropagation-based learning as applied to multilayer \acp{SNN} \cite{Gardner2015} and recurrently connected networks \cite{Brea2013,JimenezRezende2014}. Despite this, a potential drawback to this approach comes from its reliance on a stochastic neuron model for generating output spikes; although this model is well suited to reinforcement-based learning which relies on variable spiking for stochastic exploration \cite{Gardner2013}, it is less well suited to the supervised learning of precisely timed output spikes where variable responses become more of a hindrance.

To address these shortcomings, we present here two supervised learning rules, termed \acs{INST} and \acs{FILT}, which are initially derived based on the statistical method of \cite{Pfister2006}, but later adapted for compatibility with the deterministically spiking \ac{LIF} neuron model. In this way, these rules claim a stronger theoretical basis than many existing spike-based learning methods, and yet still allow for the learning of precisely timed output spikes. We then use these rules for demonstrative purposes to explore the conditions under which synaptic plasticity most effectively takes place in \acp{SNN} to allow for precise temporal encoding. These two rules differ in their formulation with respect to the treatment of output spike trains: while \ac{INST} simply relies on the \emph{instantaneous} difference between a target and actual output spike train to inform synaptic weight modifications, \ac{FILT} goes a step further, and \emph{exponentially filters} output spike trains in order to provide more stable weight changes. By this filtering mechanism, we find the \ac{FILT} rule is able to match the high performance of the E-learning \acl{CHRON} rule. We conclude by indicating the increased biological relevance of the \ac{FILT} rule over many existing spike-based supervised methods, based on this spike train filtering mechanism.

This work is organised as follows. First, the \ac{INST} and \ac{FILT} learning rules are formulated for \acp{SNN} consisting of deterministic \ac{LIF} neurons, and compared with existing, and structurally similar, spike-based learning rules. Next, synaptic weight changes triggered by the \ac{INST} and \ac{FILT} rules are analysed under various conditions, including their dependence on the relative timing difference between pre- and postsynaptic spikes, and more generally their dynamical behaviour over time. The proposed rules are then tested in terms of their accuracy when encoding large numbers of arbitrarily generated spike patterns using temporally-precise output spikes. For comparison purposes, results are also obtained for the technically efficient E-learning \acl{CHRON} rule. Finally, the rules are discussed in relation to existing supervised methods, as well as their their biological significance.

\section*{Methods}

This section proposes two supervised learning rules for \acp{SNN}, termed \ac{INST} and \ac{FILT}, that are initially formulated using the statistical approach of \cite{Pfister2006} for analytical rigour, but later adapted for use with a deterministically spiking neuron model for the purpose of precise temporal encoding.  
This section begins by describing the simplified \acl{SRM}, underpinning the formulation of the \ac{INST} and \ac{FILT} synaptic plasticity rules.

\subsection*{Single Neuron Model} \label{subsec:Neuron_model}

The \ac{LIF} neuron is a commonly used spiking neuron model, owing to its relative simplicity and analytical tractability, and represents a special case of the more general Spike Response Model \cite{Gerstner2002}. For these reasons, we begin our analysis by considering a single postsynaptic neuron $i$ with a membrane potential $u_i$ at time $t$, defined by the simplified \ac{SRM}:
\begin{equation} \label{eq:potential}
u_i(t|\mathbf{x}, y_i) := \sum_j w_{ij} \sum_{t_j^f \in x_j} \epsilon (t - t_j^f) + \sum_{t_i^f \in y_i} \kappa (t - t_i^f) \;,
\end{equation}
where the membrane potential is measured with respect to the neuron's resting potential. This equation signifies a dependence of the neuron's membrane potential on its presynaptic input pattern $\mathbf{x} = \{x_1, x_2, \ldots, x_{n_i}\}$ from $n_i$ synapses, as well as its own sequence of emitted output spikes, $y_i(t) = \{t_i^1, t_i^2, \ldots, \hat{t}_i < t\}$, where $\hat{t}_i$ is its latest spike before $t$. An actual output spike occurs at a time $t_i^f$ when $u_i$ crosses the neuron's firing threshold $\vartheta$ from below. The first term on the RHS of the above equation describes a weighted summation of the presynaptic input: the parameter $w_{ij}$ refers to the synaptic weight from a presynaptic neuron $j$, the kernel $\epsilon$ corresponds to the shape of an evoked \ac{PSP} and $x_j = \{t_j^1, t_j^2, \ldots\}$, $x_j \in \mathbf{x}$, is a list of presynaptic firing times from $j$. The second term on the RHS describes the refractoriness of the neuron due to postsynaptic spiking, controlled by the reset kernel $\kappa$. 

In more detail, the \ac{PSP} kernel evolves according to
\begin{equation} \label{eq:PSP_kernel_current}
\epsilon(s) = \frac{1}{C} \int_{s' = 0}^\infty \exp\left(-\frac{s'}{\tau_m}\right) \alpha(s - s')\, \mathrm{d}{s'}\, \Theta(s) \;,
\end{equation}
where $C$ is the neuron's membrane capacitance and $\alpha$ describes the time course of a postsynaptic current elicited due to a presynaptic spike. The term $\Theta(s)$ is the Heaviside step function, and is defined such that $\Theta(s) = 1$ for $s \geq 0$ and $\Theta(s) = 0$ otherwise. Here we approximate the postsynaptic current's time course using an exponential decay \cite{Gerstner2002}:
\begin{equation} \label{eq:LIF0_alpha_kernel}
\alpha(s) =  \frac{q}{\tau_s} \exp\left(-\frac{s}{\tau_s}\right)\, \Theta(s) \;,
\end{equation}
where $q$ is the total charge transferred due to a single presynaptic spike and $\tau_s$ is a synaptic time constant. Hence, using Eq.~(\ref{eq:LIF0_alpha_kernel}), the integral of Eq.~(\ref{eq:PSP_kernel_current}) can be evaluated to yield the \ac{PSP} kernel:
\begin{equation} \label{eq:PSP_kernel}
\epsilon(s) = \epsilon_0\, \left[ \exp\left( -\frac{s}{\tau_m} \right) - \exp\left( -\frac{s}{\tau_s} \right) \right]\, \Theta(s) \;,
\end{equation}
where its coefficient is given by $\epsilon_0 = \frac{q}{C}\, \frac{\tau_m}{\tau_m - \tau_s}$. The reset kernel in Eq.~(\ref{eq:potential}) evolves according to
\begin{equation} \label{eq:reset_kernel}
\kappa(s) = \kappa_0\, \exp\left(-\frac{s}{\tau_m}\right)\, \Theta(s) \;,
\end{equation}
with its coefficient given by $\kappa_0 = -(\vartheta - u_r)$, where the reset potential $u_r$ is the value the neuron's membrane potential is set to immediately after a postsynaptic spike is fired.

\begin{figure}[t]
\includegraphics[max width=\textwidth]{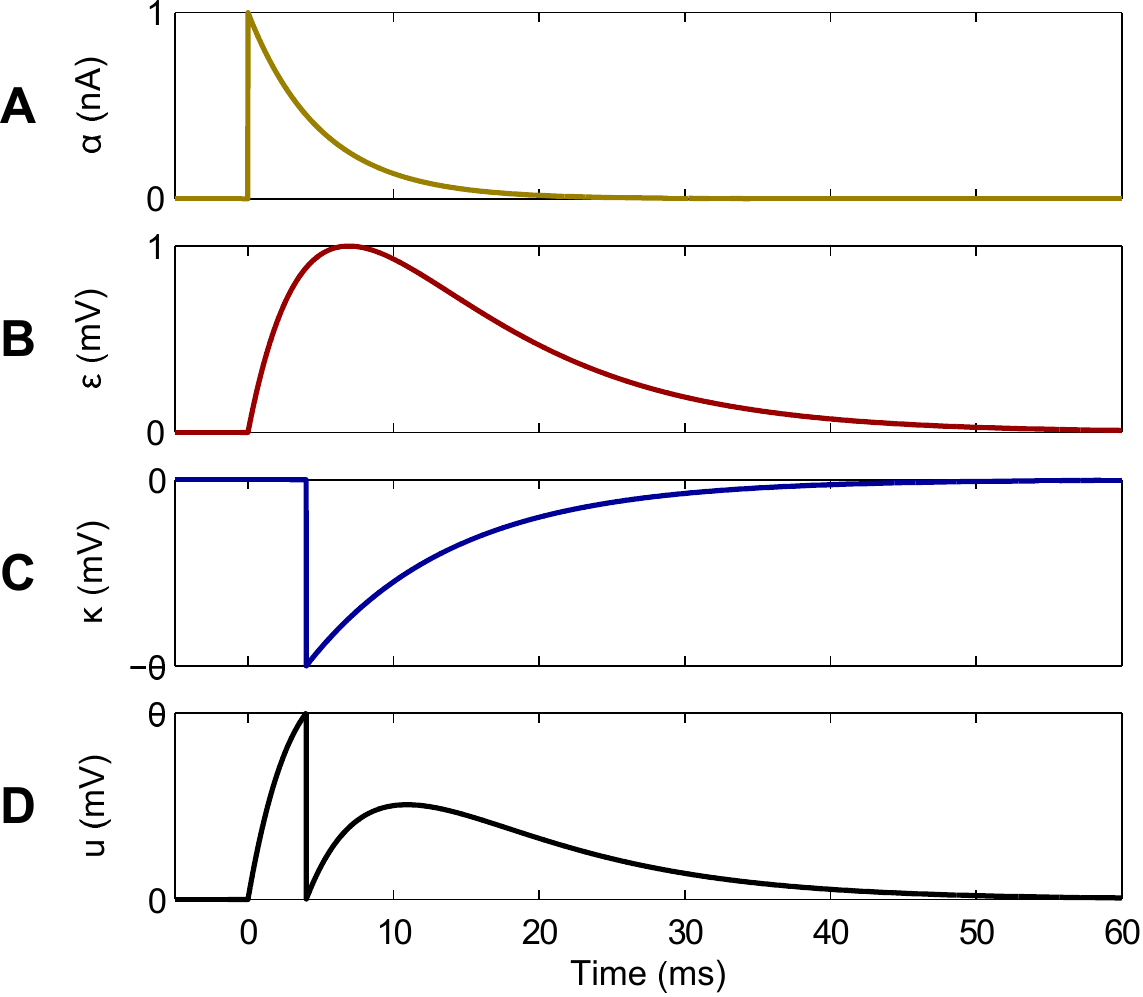}
\caption{
{\bf Illustration of the postsynaptic kernels used in this analysis, and an example of a resulting postsynaptic membrane potential.}
(A) The time course of the postsynaptic current kernel $\alpha$. (B) The \ac{PSP} kernel $\epsilon$. (C) The reset kernel $\kappa$. (D) The resulting membrane potential $u_i$ as defined by Eq.~(\ref{eq:potential}). In this example, a single presynaptic spike is received at $t_j = \SI{0}{ms}$, and a postsynaptic spike is generated at $t_i = \SI{4}{ms}$ from selectively tuning both the synaptic weight $w_{ij}$ and firing threshold $\vartheta$ values. We take $C = \SI{2.5}{nF}$ for the neuron's membrane capacitance, such that the postsynaptic current attains a maximum value of \SI{1}{nA}.
}
\label{fig1}
\end{figure}

In our analysis we set the model parameters as follows: $\epsilon_0 = \SI{4}{mV}$, $\tau_m = \SI{10}{ms}$, $\tau_s = \SI{5}{ms}$, $\vartheta = \SI{15}{mV}$ and $u_r = \SI{0}{mV}$; for these choices of parameters, a single presynaptic spike evokes a \ac{PSP} with a maximum value of \SI{1}{mV} after a lag time close to \SI{7}{ms}, and the postsynaptic neuron's membrane potential is reset to its resting value of \SI{0}{mV} immediately after firing. Shown in Fig.~\ref{fig1} are graphical illustrations of the postsynaptic current, \ac{PSP} and reset kernels, as well an example of a resulting postsynaptic membrane potential as defined by Eq.~(\ref{eq:potential}).

We now explore in more detail the spike generation mechanism of the postsynaptic neuron. Currently, firing events are considered to take place only when the neuron's membrane potential crosses a predefined firing threshold $\vartheta$. Alternatively, however, we may instead consider output spikes that are generated via a stochastic process with a time-dependent, instantaneous firing rate $\rho_i$, such that firing events may occur even at moments when the neuron's membrane potential is below the firing threshold. The instantaneous firing rate $\rho_i$ is formally referred to as the stochastic intensity of the neuron, and arbitrarily depends on the distance between the neuron's membrane potential and formal firing threshold $\vartheta$ according to
\begin{equation} \label{eq:stochastic_intensity}
\rho_i(t) = g ( u_i(t) - \vartheta ) \;,
\end{equation}
where $u_i$ is defined by Eq.~(\ref{eq:potential}) and $g$ is an arbitrary function that is commonly referred to as the neuron's `escape rate' \cite{Gerstner2002}. 

Various choices exist to define the functional form of the neuron's escape rate. A common choice is to assume an exponential dependence:
\begin{equation} \label{eq:EXP_rate}
g(u_i(t) - \vartheta) = \rho_0 \exp \left( \frac{ u_i(t) - \vartheta }{ \Delta u } \right) \;,
\end{equation}
where $\rho_0$ is the instantaneous firing rate of the neuron at threshold $\vartheta$, and the parameter $\Delta u$ determines the `smoothness' of the firing rate about the threshold \cite{Jolivet2006}. It is important to note that in taking the limit $\Delta u \rightarrow 0$ the deterministic \ac{LIF} model can be recovered, the utility of which shall become apparent later.

\subsection*{Supervised Learning Method} \label{subsec:Optimal_STDP}

Implementing a stochastic model for generating postsynaptic spikes according to Eq.~(\ref{eq:stochastic_intensity}) is advantageous, given that it allows for the determination of the likelihood of generating some desired sequence of target output spikes $y_i^{\mathrm{ref}} = \{\tilde{t}_i^1, \tilde{t}_i^2, ..., \tilde{t}_i^{n_s} \}$ containing $n_s$ spikes in response to an input spike pattern $\mathbf{x}$. As shown originally by \cite{Pfister2006}, the log-likelihood is given by
\begin{equation} \label{eq:log_PDF}
\log P( y_i^{\mathrm{ref}}|\mathbf{x} ) = \int_0^T \log \left( \rho_i(t|\mathbf{x}, y_i^\mathrm{ref}) \right) \mathcal{Y}_i^{\mathrm{ref}}(t) - \rho_i(t|\mathbf{x}, y_i^\mathrm{ref})\, \mathrm{d}t \;,
\end{equation}
where $\mathcal{Y}_i^{\mathrm{ref}}(t) = \sum_{\tilde{t}_i^f \in y_i^{\mathrm{ref}}} \delta(t - \tilde{t}_i^f)$ is a target postsynaptic spike train and $T$ is the duration over which the input pattern $\mathbf{x}$ is presented. Importantly, since the neuron model is described by the linear \ac{SRM} and the escape rate is exponential, the log-likelihood is a concave function of its parameters \cite{Paninski2004}. Log-concavity is ideal since it ensures no non-global local maxima exist in the likelihood, thereby allowing for computationally efficient parameter optimisation methods.

In our analysis, we seek to maximise the log-likelihood of a postsynaptic neuron generating a desired target output spike train $\mathcal{Y}_i^{\mathrm{ref}}$ through modifying the strengths of synaptic weights in the network. This can be achieved through the technique of gradient ascent, such that a synaptic weight $w_{ij}$ is modified according to
\begin{equation} \label{eq:grad_ascent}
\Delta w_{ij}^\mathrm{ref} \sim \frac{\partial \log P( y_i^{\mathrm{ref}}|\mathbf{x} )}{\partial w_{ij}} \;.
\end{equation}
Hence, taking the derivative of Eq.~(\ref{eq:log_PDF}) and using Eq.~(\ref{eq:potential}) provides the gradient of the log-likelihood:
\begin{equation} \label{eq:D_log_PDF}
\frac{\partial \log P( y_i^\mathrm{ref}|\mathbf{x} )}{\partial w_{ij}}  = 
\int_0^T 
\frac{ \rho_i'(t|\mathbf{x}, y_i^\mathrm{ref}) }{ \rho_i(t|\mathbf{x}, y_i^\mathrm{ref}) } \left[ \mathcal{Y}_i^{\mathrm{ref}}(t) - \rho_i(t|\mathbf{x}, y_i^\mathrm{ref}) \right] \sum_{t_j^f \in x_j} \epsilon (t - t_j^f)\, \mathrm{d}t \;,
\end{equation}
where $\rho_i'(t|\mathbf{x}, y_i^\mathrm{ref}) = \frac{\mathrm{d} g}{\mathrm{d} u}|_{u = u_i(t|\mathbf{x}, y_i^\mathrm{ref})}$. Furthermore, using Eq.~(\ref{eq:EXP_rate}) it follows that
\begin{equation} \label{eq:D_log_PDF_frac}
\frac{ \rho_i'(t|\mathbf{x}, y_i^\mathrm{ref}) }{ \rho_i(t|\mathbf{x}, y_i^\mathrm{ref}) } = \frac{1}{\Delta u} \;,
\end{equation}
which, in combination with Eqs.~(\ref{eq:grad_ascent}), (\ref{eq:D_log_PDF}) and (\ref{eq:D_log_PDF_frac}), provides the weight update rule:
\begin{equation} \label{eq:w_update_stoch}
\Delta w_{ij}^\mathrm{ref} = \frac{\eta}{\Delta u} \int_0^T \left[ \mathcal{Y}_i^{\mathrm{ref}}(t) - \rho_i(t|\mathbf{x}, y_i^\mathrm{ref}) \right] \sum_{t_j^f \in x_j} \epsilon (t - t_j^f)\, \mathrm{d}t \;,
\end{equation}
where $\eta$ is the learning rate. The above has been derived by \cite{Pfister2006}, and has been shown to well approximate results of experimental protocols on synaptic plasticity which depend on the coincidence of pre- and postsynaptic firing times \cite{Bi1998}.

In our approach, however, we wish to instead consider a learning rule that depends on the intrinsic dynamics of a postsynaptic neuron, rather than artificially clamping its firing activity to its target response. To this end, we adjust the weight update rule of Eq.~(\ref{eq:w_update_stoch}) to the following rule:
\begin{equation} \label{eq:w_update_intrinsic}
\Delta w_{ij} = \frac{\eta}{\Delta u} \int_0^T \left[ \mathcal{Y}_i^{\mathrm{ref}}(t) - \rho_i(t|\mathbf{x}, y_i) \right] \sum_{t_j^f \in x_j} \epsilon (t - t_j^f)\, \mathrm{d}t \;,
\end{equation}
where we have substituted $\rho_i(t|\mathbf{x}, y_i^\mathrm{ref})$ with $\rho_i(t|\mathbf{x}, y_i)$, such that the instantaneous firing rate of the postsynaptic neuron depends on its actual sequence of emitted output spikes $y_i$ rather than its target output $y_i^\mathrm{ref}$.
Although Eq.~(\ref{eq:w_update_intrinsic}) is an approximation of the theoretical result of Eq.~(\ref{eq:w_update_stoch}), it can be shown that it nevertheless converges towards a similar solution when certain conditions are satisfied, depending on the magnitude of $\Delta u$ and $\kappa_0$, and the relative timing displacements between target and actual output spikes (see \nameref{app}).

\subsection*{INSTantaneous-error (INST) Synaptic Plasticity Rule} \label{subsec:INST_rule}

The weight update rule of Eq.~(\ref{eq:w_update_stoch}) has been derived by taking a maximum-likelihood approach based on a stochastic spiking neuron model, but can be adapted, using Eq.~\eqref{eq:w_update_intrinsic}, to the case of a deterministically firing \ac{LIF} neuron model to allow for precise temporal encoding. Specifically, if the limit $\Delta u \rightarrow 0$ is taken for the stochastic threshold parameter in Eq.~(\ref{eq:EXP_rate}), the stochastic intensity of an intrinsically spiking neuron instead assumes one of two values:
\begin{equation} \label{eq:EXP_rate_deterministic}
\rho_i(t) = 
  \begin{cases}
   \delta(t - t_i^f) & \text{for } u_i(t_i^f) > \vartheta \\
   0 & \text{otherwise} \;,
  \end{cases}
\end{equation}
where the term $\delta(t - t_i^f)$ is the Dirac delta distribution about an actual postsynaptic firing time $t_i^f \in y_i$, since immediately after a spike is emitted: $u_i(t_i^{f+}) < \vartheta$ as a result of the reset term in Eq.~(\ref{eq:potential}). In this way, the postsynaptic neuron's stochastic intensity can be substituted with its output spike train $\rho_i(t) \rightarrow \mathcal{Y}_i(t)$, where $\mathcal{Y}_i(t) = \sum_{t_i^f \in y_i} \delta(t - t_i^f)$. Hence, using Eq.~\eqref{eq:w_update_intrinsic} and the result for Eq.~\eqref{eq:EXP_rate_deterministic}, a deterministic adaptation of Eq.~(\ref{eq:w_update_stoch}) is given by
\begin{equation} \label{eq:w_update_INST}
\lim_{\Delta u \rightarrow 0} \Delta w_{ij} = \eta \int_0^T \left[ \mathcal{Y}_i^{\mathrm{ref}}(t) - \mathcal{Y}_i(t) \right] \sum_{t_j^f \in x_j} \epsilon (t - t_j^f)\, \mathrm{d}t \;,
\end{equation}
where we have renormalised the above equation by redefining the learning rate to maintain a finite value: $\eta \leftarrow \frac{\eta}{\Delta u}$.
Strictly speaking, taking the limit $\Delta u \rightarrow 0$ cannot be guaranteed to provide convergence towards an optimal synaptic weight solution, as is otherwise predicted for small postsynaptic timing displacements and finite $\Delta u$, but suitably simplifies the learning rule for the case of a deterministic, and intrinsically spiking, neuron model. The convergence of this simplified rule shall be experimentally analysed in detail in the Results section.
Furthermore, performing the straightforward integration of Eq.~(\ref{eq:w_update_INST}) provides the batch weight update rule:
\begin{equation} \label{eq:w_update_INST_batch}
\Delta w_{ij}^{\mathrm{INST}} = \eta \Bigg[ \sum_{\tilde{t}_i^g \in y_i^{\mathrm{ref}}} \, \sum_{t_j^f \in x_j} \epsilon (\tilde{t}_i^g - t_j^f) - \sum_{t_i^h \in y_i} \, \sum_{t_j^f \in x_j} \epsilon (t_i^h - t_j^f) \Bigg] \;,
\end{equation}
which we term the \acf{INST} synaptic plasticity rule, to reflect the discontinuous nature of the postsynaptic error signal. The \ac{INST} rule can be summarised as a two-factor learning rule: presynaptic activity describing a stimulus (first learning factor) is combined with a postsynaptic error signal (second learning factor) to elicit a final synaptic weight change.

Broadly speaking, the \ac{INST} rule falls into a class of learning rules for \acp{SNN} which depend on an instantaneous error signal to drive synaptic weight modifications. Key examples include the \ac{PSD} plasticity rule proposed in \cite{Yu2013}, the I-learning variant of the Chronotron \cite{Florian2012} and the \ac{FP} algorithm \cite{Memmesheimer2014}. Despite this, certain differences exist between \ac{INST} and the aforementioned examples. Specifically, weight updates for both \ac{PSD} and I-learning rely on the postsynaptic current $\alpha$, rather than the \ac{PSP} $\epsilon$ as is used here, as a presynaptic learning factor (compare Eqs.~(\ref{eq:LIF0_alpha_kernel}) and (\ref{eq:PSP_kernel}), respectively). The selection of $\alpha$ as a presynaptic learning factor is somewhat arbitrary, while $\epsilon$ is theoretically supported \cite{Pfister2006,Xu2013}.

Although \ac{INST} and the \ac{FP} algorithm share $\epsilon$ as their presynaptic learning factor, the \ac{FP} algorithm just takes into account the first occurrence of an error due to a misplaced postsynaptic spike, rather than accumulating all postsynaptic spike errors as for \ac{INST}. The authors' decision to restrict \ac{FP} learning to the first error in each trial was motivated by a desire to avoid non-linear accumulation of errors arising from interacting postsynaptic spikes, due to the neuron's reset term, in order that weight updates alter the future time course of the neuron's activity in a more predictable manner \cite{Memmesheimer2014}. Here we relax this constraint for the sake of biological plausibility and ease of implementation, but still ensure that target postsynaptic spikes are sufficiently separated from each other to reduce this error accumulation effect.

\subsection*{FILTered-error (FILT) Synaptic Plasticity Rule} \label{subsec:FILT_rule}

As it currently stands, the time course of the synaptic weight change $\Delta w_{ij}(t)$ resulting from Eq.~(\ref{eq:w_update_INST}) depends on the instantaneous difference between two spike trains $\mathcal{Y}_i^{\mathrm{ref}}$ and $\mathcal{Y}_i$ during learning. In other words, candidate weight updates are only effected at the precise moments in time when target or actual postsynaptic spikes are present. Although this leads to the simplified batch weight update rule of Eq.~(\ref{eq:w_update_INST_batch}), there are two distinct disadvantages to this approach. The first concerns the convergence of actual postsynaptic spikes towards matching their desired target timings;
as we shall show in the Results section, and as previously indicated in \cite{Florian2012,Memmesheimer2014}, if the temporal proximity of postsynaptic spikes is not accounted for by the learning rule, then fluctuations in the synaptic weights can emerge as a result of unstable learning. It then becomes problematic for the network to smoothly converge towards a desired weight solution, and maintain fixed output firing activity.
Secondly, from a biological standpoint it is implausible to assume that synaptic weights can be effected instantaneously at the precise timings of postsynaptic spikes. More realistically, it can be supposed that postsynaptic spikes would leave some form of synaptic trace that persists on the order of the membrane time constant, which, in combination with coincident presynaptic spiking as detected via evoked \acp{PSP}, would inform more gradual synaptic weight changes.

To address these limitations of instantaneous-error based learning we convolve the target and actual output spike trains of the postsynaptic neuron of Eq.~(\ref{eq:w_update_INST}) with an exponential kernel, thereby providing the following learning rule:
\begin{equation} \label{eq:w_update_conv}
\Delta w_{ij} = \eta \int_0^\infty \left[ \tilde{\mathcal{Y}}_i^{\mathrm{ref}}(t) - \tilde{\mathcal{Y}}_i(t) \right] \sum_{t_j^f \in x_j} \epsilon (t - t_j^f)\, \mathrm{d}t \;,
\end{equation}
where a convolved actual output spike train is equivalent to
\begin{equation} \label{eq:convolved_spike_train}
\tilde{\mathcal{Y}}_i(t) \equiv \frac{1}{\tau_q} \int_0^t \mathcal{Y}_i(t') \exp\left( -\frac{t - t'}{\tau_q} \right)  \, \mathrm{d}t' \;,
\end{equation}
and a similar equivalence for a target output spike train $\tilde{\mathcal{Y}}^{\mathrm{ref}}_i$. The decay time constant is set to $\tau_q = \SI{10}{ms}$, similar to the membrane time constant $\tau_m$, which has been indicated to give optimal performance from preliminary parameter sweeps. The upper limit of $\infty$ in Eq.~(\ref{eq:w_update_conv}) is necessary in order to account for the entire time course of convolved postsynaptic traces. Performing the integration of Eq.~(\ref{eq:w_update_conv}) using the \ac{PSP} kernel given by Eq.~(\ref{eq:PSP_kernel}) yields the batch weight update rule:
\begin{equation} \label{eq:w_update_FILT}
\Delta w_{ij}^{\mathrm{FILT}} = \eta \Bigg[ \sum_{\tilde{t}_i^g \in y_i^{\mathrm{ref}}} \, \sum_{t_j^f \in x_j} \lambda (\tilde{t}_i^g - t_j^f) - \sum_{t_i^h \in y_i} \, \sum_{t_j^f \in x_j} \lambda (t_i^h - t_j^f) \Bigg] \;,
\end{equation}
where the learning window $\lambda$ arises from interacting pre- and postsynaptic spikes, and is given by
\begin{equation} \label{eq:PSP_FILT}
\lambda(s) = 
  \begin{cases}
    \epsilon_0\, \left[ \mathcal{C}_m\, \exp\left( -\frac{s}{\tau_m} \right) - \mathcal{C}_s\, \exp\left( -\frac{s}{\tau_s} \right) \right] & \text{for } s > 0 \\
    \epsilon_0\, \left( \mathcal{C}_m - \mathcal{C}_s \right)\, \exp \left( \frac{s}{\tau_q} \right) & \text{for } s \leq 0 \;.
  \end{cases}
\end{equation}
In the above equation, the membrane and synaptic coefficient terms are $\mathcal{C}_m = \frac{\tau_m}{\tau_m + \tau_q}$ and $\mathcal{C}_s = \frac{\tau_s}{\tau_s + \tau_q}$, respectively. We term Eq.~(\ref{eq:w_update_FILT}) the \acf{FILT} synaptic plasticity rule, that depends on the smoothed difference between filtered target and actual output spike trains.

The \ac{FILT} rule falls into a class of learning rules for \acp{SNN} which rely on a smoothed error signal for weight updates, and which more effectively take into account the temporal proximity of neighbouring target and actual postsynaptic spikes, such as the \ac{SPAN} rule \cite{Mohemmed2012} and E-learning variant of the \acl{CHRON} \cite{Florian2012}. In particular, Eq.~(\ref{eq:w_update_conv}) bears a similarity with \ac{SPAN}, in the sense that weight updates depend on convolved pre- and postsynaptic spike trains. However, as for the \ac{PSD} rule, \ac{SPAN} makes no prediction for the choice of kernel function with which to convolve presynaptic spike trains. In our analysis, presynaptic spikes are suitably convolved with the \ac{PSP} kernel of Eq.~(\ref{eq:PSP_kernel}).
The exponential filtering of postsynaptic spike trains by the \ac{FILT} rule may appear arbitrary, but it is not unreasonable to suppose that this operation is carried out via changes in the neuron's membrane potential in response to postsynaptic spikes: especially since the filter time constant has an optimal value, as determined through preliminary parameter sweeps, that is similar to the neuron's membrane time constant, $\tau_q \approx \tau_m$.

Additionally, selecting an exponential filter simplifies the resulting batch weight update rule, and coincidentally provides a resemblance of \ac{FILT} to the \acf{vRD} as is used to measure the (dis)similarity between neuronal spike trains \cite{Rossum2001}, where the \ac{vRD} is defined by
\begin{equation} \label{eq:vRD}
	\mathcal{D}(\mathcal{Y}_i, \mathcal{Y}_i^\mathrm{ref}) := \frac{1}{\tau_q} \int_0^{\infty} [\tilde{\mathcal{Y}}_i(t) - \tilde{\mathcal{Y}}_i^\mathrm{ref}(t)]^2 \mathrm{d}t \;.
\end{equation}
Hence, the \ac{FILT} rule might also be interpreted as an error minimisation procedure that reduces, by gradient descent, a \ac{vRD}-like error function measuring the distance between target and actual postsynaptic spike trains.

\section*{Results}

\subsection*{Analysis of the Learning Rules} \label{sec:Plasticity_analysis}

We first analyse the validity of synaptic weight modifications resulting from the \ac{INST} and \ac{FILT} rules under general learning conditions. For ease of analysis we examine just the weight change between a single pair of pre- and postsynaptic neurons: each emitting a single spike at times $t_j$ and $t_i$, respectively. A single target output spike at time $\tilde{t}_i$ is also imposed, which the postsynaptic neuron must learn to match. 

This subsection is organised as follows. First, simplified weight update rules for \ac{INST} and \ac{FILT} are presented based on single pre- and postsynaptic spiking. Next, two distinct scenarios of weight change driven by each learning rule are examined. The first scenario examines the weight change as a function of the relative timing difference between a target postsynaptic spike and presynaptic spike. The second scenario then considers the dynamics of each learning rule by examining their weight change as a function of the current weight value, with the intent of establishing their potential for stable convergence towards a desired weight solution.

\paragraph{Synaptic weight updates for single spikes.}
According to the definition of the \ac{INST} rule in Eq.~(\ref{eq:w_update_INST_batch}), the synaptic weight change triggered by single spikes is given by
\begin{equation} \label{eq:INST_spike}
\Delta w_{ij}^{\mathrm{INST}} = \eta \left[ \epsilon (\tilde{t}_i - t_j) - \epsilon (t_i - t_j) \right] \;,
\end{equation}
that is simply the difference between two \ac{PSP} kernels. For the above equation there exist several conditions under which no weight change results, including the trivial case when both \ac{PSP} terms are equal to zero as a result of post- before presynaptic spiking (i.e. $t_i, \tilde{t}_i \leq t_j$). Additionally, no weight change occurs when both \ac{PSP} terms share the same value: ideally this would take place when target and actual output spikes become aligned, i.e. when $t_i = \tilde{t}_i$. However, no weight change is also possible for non-aligned output spikes, since the \ac{PSP} kernel assumes the same value for two distinct lag times (compare the rising and falling segments of the \ac{PSP} curve in Fig.~\ref{fig1}B).

\begin{figure}[t]
\includegraphics[max width=\textwidth]{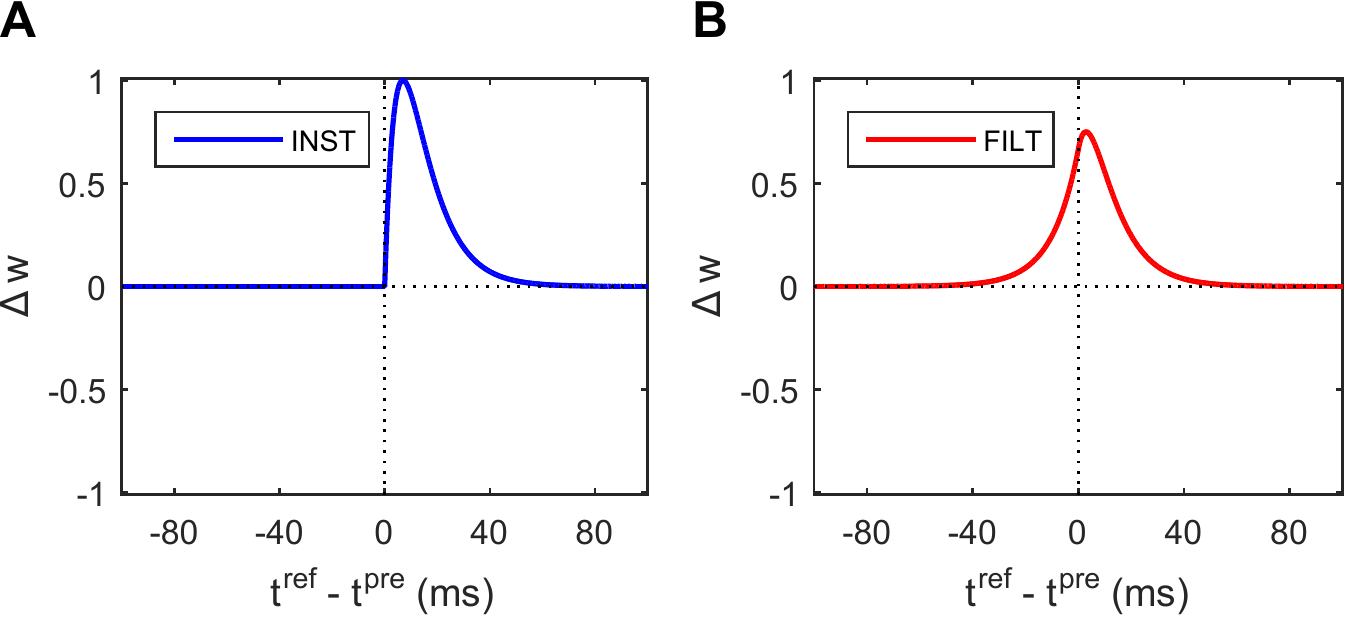}
\caption{
{\bf Dependence of synaptic weight change $\Delta w$ on the relative timing difference between a target postsynaptic spike and input presynaptic spike: $t^{\mathrm{ref}}$ and $t^{\mathrm{pre}}$, respectively.} (A) Leaning window of the \ac{INST} rule. (B) Learning window of the \ac{FILT} rule. The peak $\Delta w$ values for \ac{INST} and \ac{FILT} correspond to relative timings of just under \SIlist[list-units = single]{7; 3}{ms}, respectively. Both panels show the weight change in the absence of an actual postsynaptic spike.
}
\label{fig2}
\end{figure}

Similarly, the \ac{FILT} batch weight update rule of Eq.~(\ref{eq:w_update_FILT}) can be simplified for single pre- and postsynaptic spikes:
\begin{equation} \label{eq:FILT_spike}
\Delta w_{ij}^{\mathrm{FILT}} = \eta \left[ \lambda (\tilde{t}_i - t_j) - \lambda (t_i - t_j) \right] \;,
\end{equation}
that is the difference between two synaptic learning windows, $\lambda$, as defined by Eq.~(\ref{eq:PSP_FILT}). As with the \ac{INST} rule, there is no weight change for the above equation in the event that both $\lambda$ terms share the same value: like the \ac{PSP} kernel $\epsilon$ there exists two distinct lag times for which $\lambda$ assumes the same value (see the form of the curve in Fig.~\ref{fig2}B), hence target and actual postsynaptic spikes need not necessarily be aligned to elicit a zero weight change. Unlike the \ac{INST} rule, however, a weight change for \ac{FILT} can be non-zero for post- before presynaptic spiking.

In the rest of this subsection we start by simply examining the synaptic weight change as a function of the relative timing difference between a target postsynaptic spike and input presynaptic spike, in the absence of an actual postsynaptic spike, in order to establish the temporal learning window of each synaptic plasticity rule. We then graphically study the dynamics of each rule by plotting their phase space diagrams, to predict their long term temporal evolution of the synaptic weight towards a limiting value. For demonstrative purposes the learning rate of the \ac{INST} and \ac{FILT} rule is set to unity here, although there is no qualitative change in the results for different values.

\paragraph{Temporal window of the learning rules.}
Shown in Fig.~\ref{fig2} is the synaptic weight change for each learning rule as a function of the relative timing difference between a target postsynaptic spike and presynaptic spike, denoted by $t^{\mathrm{ref}} - t^{\mathrm{pre}}$, including for negative relative timings. Both panels in this figure correspond to the absence of an actual postsynaptic spike, to clearly illustrate the temporal locality of each synaptic plasticity rule.

From the top panel of Fig.~\ref{fig2}A for the \ac{INST} rule, it is observed that the plot of the synaptic change simply follows the form of a \ac{PSP} kernel. In this case, the synaptic change is zero for negative values of the relative timing difference, demonstrating the causality of a presynaptic spike in eliciting a desired postsynaptic spike. Interestingly, the top panel of Fig.~\ref{fig2}B for the \ac{FILT} rule instead demonstrates a more symmetrical dependence of synaptic change on the relative timing difference, which is centred just right of the origin. This contrasts with the \ac{INST} rule, and can be explained by the \ac{FILT} rule instead working to minimise the \emph{smoothed} difference between a target and actual spike train, rather than just their \emph{instantaneous} difference; in other words, even if an actual, emitted postsynaptic spike cannot technically be aligned with its target, then a close match is deemed to be sufficient under \ac{FILT}.

\paragraph{Dynamics of the learning rules.}
We now turn to examining the dynamics of synaptic weight changes elicited under the \ac{INST} and \ac{FILT} learning rules, in order to predict their long term behaviour. 

At this point it is necessary to discuss the relationship between the timing of an actual output spike fired by a postsynaptic neuron and the shape of a \ac{PSP} evoked by an input spike. In response to a single synapse, a postsynaptic neuron is constrained to firing an output spike with a lag time up to the peak value of the \ac{PSP} kernel, since this is the only region over which the neuron's membrane potential can be adjusted to cross its firing threshold from below. For this reason, we confine our analysis here to examining the dynamics of synaptic weight changes arising from postsynaptic spikes that occur over the rising segment of the \ac{PSP} curve, corresponding to lag times up to $\sim \SI{7}{ms}$ for our choice of parameters, as is visualised in Fig.~\ref{fig1}B.

In more detail, if a postsynaptic neuron $i$ receives a single input spike at $t_j = \SI{0}{ms}$ from a synapse $j$ with weight $w_{ij} \geq \frac{\vartheta}{\epsilon^\mathrm{peak}}$, then its actual output firing time $t_i$ is provided by the relation:
\begin{equation} \label{eq:dynamics1}
w_{ij}\, \epsilon(t_i) = \vartheta \;,
\end{equation}
where the conditional parameter $\epsilon^\mathrm{peak}$ corresponds to the maximum value $\epsilon$ attains after a lag time of $s^{\mathrm{peak}} = \frac{\tau_m \tau_s}{\tau_m - \tau_s} \log \left( \frac{\tau_m}{\tau_s} \right)$. For values $w_{ij} < \frac{\epsilon^\mathrm{peak}}{\vartheta}$ there is insufficient synaptic drive to initiate an output spike. Furthermore, if we isolate $\epsilon$ over its sub-domain: $[0, s^\mathrm{peak}]$, corresponding to the rising segment of the \ac{PSP}, then the actual output firing time can be explicitly written in terms of an inverse function of $\epsilon$:
\begin{equation} \label{eq:dynamics2}
t_i = \epsilon^{-1} \left( \frac{\vartheta}{w_{ij}} \right) \;,
\end{equation}
and which can be determined as
\begin{equation} \label{eq:dynamics3}
t_i = \tau_m\, \log \left( \frac{2}{1 + \sqrt{1 - \frac{4\, \vartheta}{\epsilon_0\, w_{ij}}}} \right) \;,
\end{equation}
when assuming the proportionality between the membrane and synaptic time constants: $\tau_s = \frac{\tau_m}{2}$. As described by the above equation, an increase in the synaptic weight works to shift an actual spike backwards in time, and a decrease in the synaptic weight shifts an actual spike forwards in time. By this process, a neuron can be trained to find a desirable synaptic weight value which minimises the temporal difference of an actual output spike with respect to its target. 

\begin{figure}[t]
\includegraphics[scale=0.95]{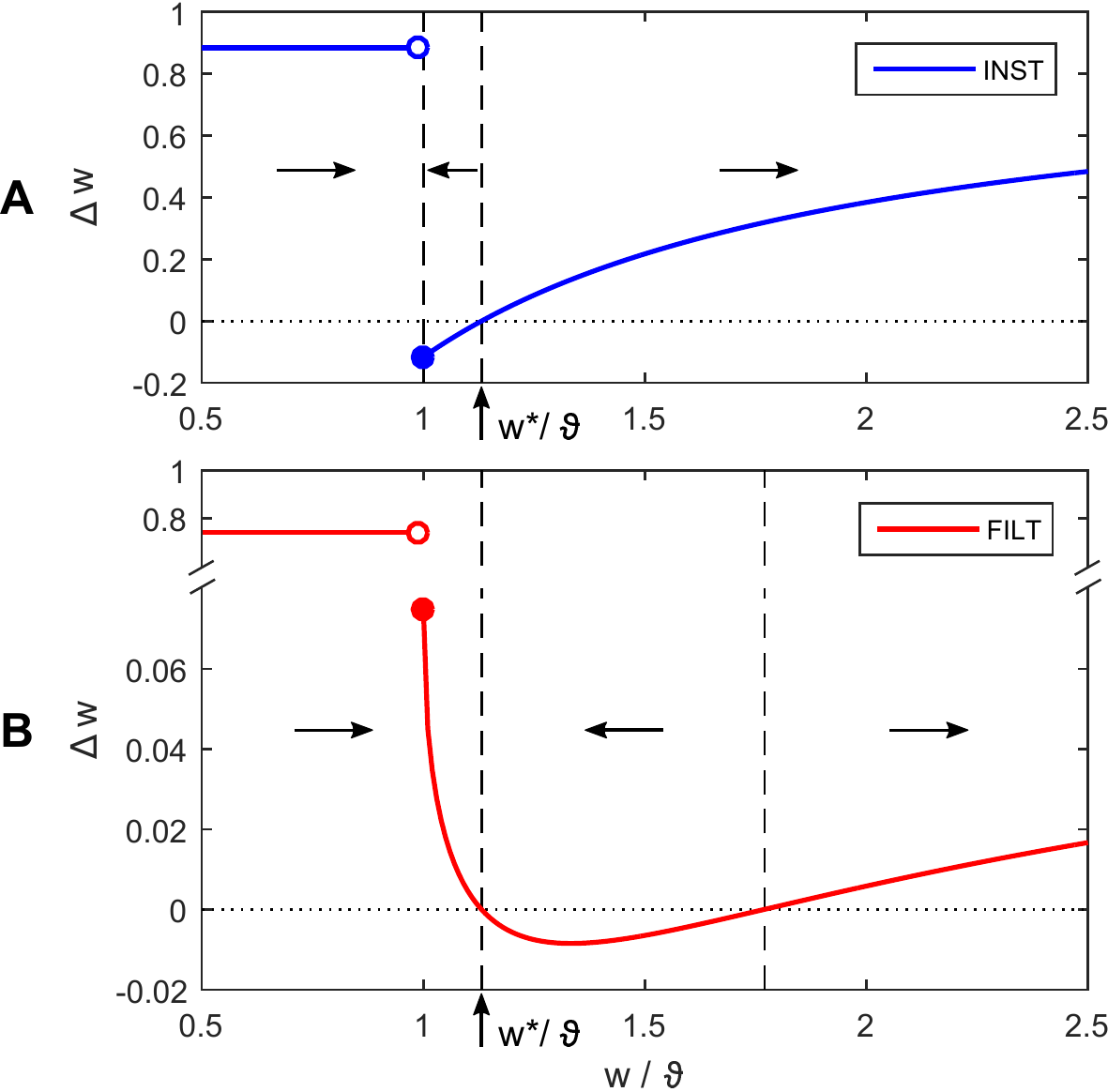}
\caption{
{\bf Phase portraits of the \acs{INST} and \acs{FILT} synaptic plasticity rules for a single synapse, each plotting the change in the synaptic weight $\Delta w$ as a function of its current strength relative to threshold $w / \vartheta$.} In this example, a postsynaptic neuron receives an input spike at time $t^\mathrm{pre} = \SI{0}{ms}$ from a single synapse with weight $w$. The postsynaptic neuron must learn to match a target output spike time $t^\mathrm{ref} = \SI{4}{ms}$, which corresponds to a desired synaptic weight solution $w^*$ as indicated in both panels. The actual output spike fired by the neuron is shifted backwards in time for positive $\Delta w$, and vice versa for negative $\Delta w$. The horizontal arrows in each panel show the direction in which $w$ evolves, and are separated by the vertical dashed lines. The peak \ac{PSP} value $\epsilon^\mathrm{peak} = \SI{1}{mV}$ (see Methods) results in an actual output spike being fired for $w / \vartheta \geq 1$.
}
\label{fig3}
\end{figure}

Using Eqs.~(\ref{eq:dynamics2}) and (\ref{eq:INST_spike}), assuming $t_j = \SI{0}{ms}$ and taking $\eta = 1$, the \ac{INST} weight update rule can be summarised as follows for a single synapse:
\begin{equation} \label{eq:INST_dynamics}
\Delta w_{ij}^{\mathrm{INST}} = 
  \begin{cases}
    \epsilon (\tilde{t}_i) - \frac{\vartheta}{w_{ij}} & \text{for } w_{ij} \geq \frac{\vartheta}{\epsilon^\mathrm{peak}} \\
    \epsilon (\tilde{t}_i) & \text{for } w_{ij} < \frac{\vartheta}{\epsilon^\mathrm{peak}} \;,
  \end{cases}
\end{equation}
that is a discontinuous function of the synaptic strength $w_{ij}$. The above synaptic plasticity rule is plotted as a phase portrait in Fig.~\ref{fig3}A, illustrating the change in the synaptic weight as a function of its current strength. This figure  displays two states of the postsynaptic neuron: the first of which is quiescence for subthreshold weight values, and the other firing activity for suprathreshold values. The sudden transition from positive to negative $\Delta w$ about $w / \vartheta = 1$ (coinciding with $\epsilon^\mathrm{peak} = \SI{1}{mV}$) corresponds to a transition between these two states, whereupon the neuron first responds with an output spike. This transition point also acts as an attractor for the system, to the extent that weight values $-\infty < w < w^*$ are drawn towards it, where $w^*$ is a desired weight solution. This point is unstable, however, due to the discontinuity in $\Delta w$, and ultimately results in fluctuations of $w$. This unstable attractor is detrimental to network performance for two key reasons: the first being that it potentially draws $w$ away from its target value of $w^*$, and the second arising from its tendency to drive variable postsynaptic firing activity as the neuron is effectively `switched on and off' due to fluctuations in $w$ about $\vartheta$. The second fixed point in Fig.~\ref{fig3}A, indicated by the second dashed line from the left, is a repeller, and, unless $w$ is exactly equal to $w^*$, will work to repel $w$. This point in particular leads us to predict that learning is unlikely to precisely converge under the \ac{INST} rule, and especially for large initial values of $w$ for which divergence will result.

Starting with Eq.~(\ref{eq:FILT_spike}) and again assuming: $t_j = \SI{0}{ms}$ and $\eta = 1$, the \ac{FILT} rule can be summarised as follows for a single synapse:
\begin{equation} \label{eq:FILT_dynamics}
\Delta w_{ij}^{\mathrm{FILT}} = 
  \begin{cases}
    \lambda (\tilde{t}_i) - \lambda (t_i) & \text{for } w_{ij} \geq \frac{\vartheta}{\epsilon^\mathrm{peak}} \\
    \lambda (\tilde{t}_i) & \text{for } w_{ij} < \frac{\vartheta}{\epsilon^\mathrm{peak}} \;,
  \end{cases}
\end{equation}
where the actual output firing time $t_i$, emitted over the \ac{PSP}'s rising segment, is determined by Eq.~(\ref{eq:dynamics3}). Shown in Fig.~\ref{fig3}B is a phase portrait of the \ac{FILT} rule, where the weight change is plotted as a function of its current strength. Similarly as discussed before in relation to the \ac{INST} rule, the postsynaptic neuron here exhibits two distinct states: quiescence for $w < \vartheta$, and firing activity for $w \geq \vartheta$. As for the \ac{INST} rule there is a discontinuity in $\Delta w$ as the neuron crosses its firing threshold, however, unlike with \ac{INST}, $\Delta w$ remains positive until the desired weight solution is reached at $w^*$. This has the effect of shifting the system attractor to $w^*$ (indicated by first dashed line from the left), as well as making it a stable point by avoiding a discontinuous change in $\Delta w$. Conversely, the second dashed line corresponds to an unstable fixed point, and works to repel $w$. Taken together, it follows that for sufficiently small initial values of $w$ the \ac{FILT} rule predictably leads to convergence in learning, with a stable synaptic weight solution.
\begin{figure}[t]
\includegraphics[max width=\textwidth]{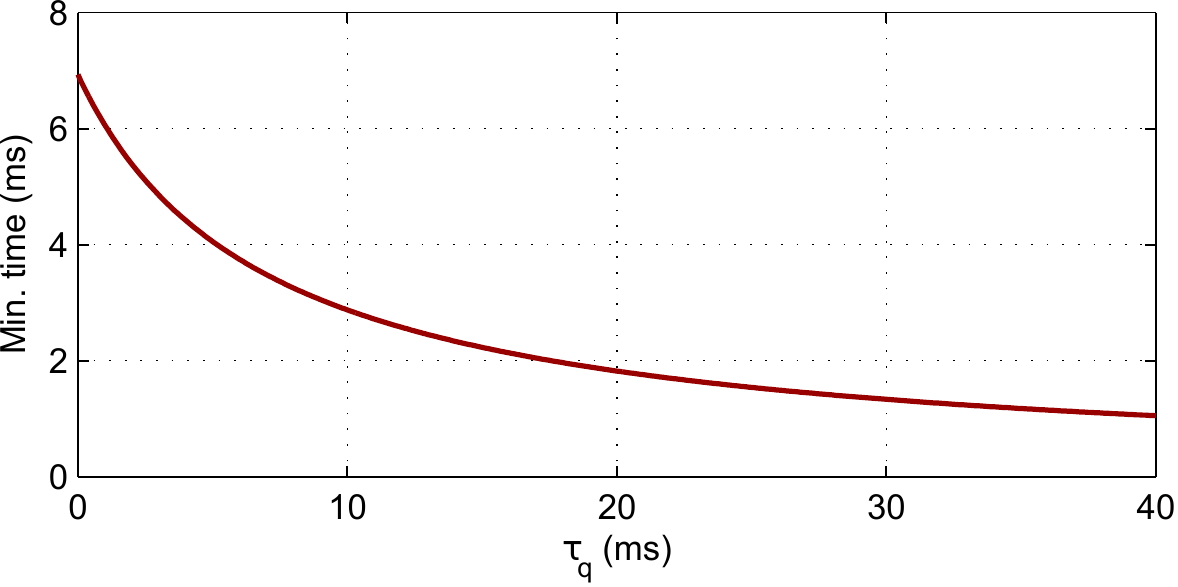}
\caption{
{\bf The minimum target output firing time $\tilde{t}_i^\mathrm{min}$, relative to an input spike time, that can accurately be learned using the \ac{FILT} rule, plotted as a function of the filter time constant $\tau_q$.} This figure makes predictions based on a single synapse with an input spike at \SI{0}{ms}. At $\tau_q = \SI{0}{ms}$ the minimum time $\tilde{t}_i^\mathrm{min}$ is equivalent to $s^\mathrm{peak}$, that is the lag time corresponding to the maximum value of the \ac{PSP} kernel, and \ac{FILT} becomes equivalent to \ac{INST}. As a reference, the value $\tau_q = \SI{10}{ms}$ was selected for use in our computer simulations, which was indicated to give optimal performance on preliminary runs.
}
\label{fig4}
\end{figure}

However, depending on the filter time constant $\tau_q$, there is a limit on the minimum value of $\tilde{t}_i$ that can reliably be learned when using the \ac{FILT} rule. In more detail, and from using Eqs.~(\ref{eq:FILT_dynamics}) and (\ref{eq:PSP_FILT}), it can be shown that this lower bound on $\tilde{t}_i$ for stable convergence of the learning rule is given by
\begin{equation} \label{eq:FILT_dynamics2}
\tilde{t}_i^\mathrm{min} = \frac{\tau_m \tau_s}{\tau_m - \tau_s} \log \left( \frac{\tau_m + \tau_q}{\tau_s + \tau_q} \right) \;,
\end{equation}
corresponding to the moment at which the associated weight solution $w^*$ changes from a stable to unstable fixed point. In other words, values of $\tilde{t}_i < \tilde{t}_i^\mathrm{min}$ would result in diminished learning, as $w$ is instead repelled away from its target value of $w^*$. From Eq.~(\ref{eq:FILT_dynamics2}) it is clear that the free parameter $\tau_q$ influences the stability of the learning rule, such that $\tau_q \in [0, \infty)$ is mapped to a minimum target timing of $\tilde{t}_i^\mathrm{min} \in [s^{\mathrm{peak}}, 0)$ as illustrated in Fig.~\ref{fig4} for $\tau_q \leq \SI{40}{ms}$. Therefore, decreasing $\tilde{t}_i^\mathrm{min}$ with respect to its parameter $\tau_q$ should predictably lead to increased temporal precision of the \ac{FILT} rule. As stated in the previous section, we select $\tau_q = \SI{10}{ms}$ for use in our simulations: this corresponds to a value of $\tilde{t}_i^\mathrm{min}$ that is just under \SI{3}{ms}.

\paragraph{Summary.}
This subsection has analysed synaptic weight modifications driven by the \ac{INST} and \ac{FILT} learning rules, based on single pre- and postsynaptic spiking for a single synapse. In particular, \ac{FILT} is predicted to provide convergence towards a stable and accurate solution in most cases, which depends crucially on the magnitude of its filter time constant $\tau_q$. By contrast, the \ac{INST} rule is predicted to give rise to less accurate solutions, and typically result in variable firing activity due to fluctuations in the synaptic strength close to the postsynaptic neuron's firing threshold. In fact, this instability is indicative of a key difference between the \ac{INST} rule and Pfister's learning rule as defined by Eq.~\eqref{eq:w_update_stoch}: while postsynaptic spiking, post-training, under Pfister's rule would fluctuate around its target timing, \ac{INST} would instead lead to fluctuating spikes around a timing coinciding with the peak value of the \ac{PSP}, independent of the target time. Finally, it is noted that, for analytical tractability, these dynamical predictions for \ac{INST} and \ac{FILT} have been made for single, rather than multiple, synapses. Hence, it shall be the aim of the next section to explore the validity of these learning rules in larger network sizes through numerical simulation.

\subsection*{Simulations} \label{sec:Simulations}

This subsection presents results from computer simulations testing the performance of the \ac{INST}, \ac{FILT} and E-learning rules. E-learning, henceforth referred to here as \ac{CHRON}, is used in our simulations, being an ideal benchmark against which our derived rules can be compared; \ac{CHRON} is ideal since it incorporates a mechanism for linking together target and actual postsynaptic spikes, analogous to the proposed \ac{FILT} rule in the sense that it accounts for the temporal proximity of neighbouring postsynaptic spikes, as well as allowing for a very high network capacity in terms of the maximum number of input patterns it can learn to memorise \cite{Florian2012}. It is worth noting that these three learning rules are essentially based on distinct spike train error measures: the \ac{INST} rule simply based on a momentary spike count error, the \ac{FILT} rule based on a smoothed \acl{vRD}-like error function \cite{Rossum2001}, and the \ac{CHRON} rule based on an adaptation of the \acl{VPD} measure \cite{Victor1996}.

\paragraph{Network setup.}
In simulations, the network consisted of either one or multiple postsynaptic neurons receiving input spikes from a variable number $n_i$ of presynaptic neurons in a feedforward manner. The dynamics of the postsynaptic neuron's membrane potential $u_i$ was governed according to the \ac{SRM} defined by Eq.~(\ref{eq:potential}), and output spikes were instantly generated when the neuron's membrane potential reached the formal firing threshold $\vartheta$; hence, we implemented a deterministic adaptation of the stochastic neuron model presented in Eq.~(\ref{eq:EXP_rate}), as necessitated by the derived \ac{INST} and \ac{FILT} learning rules. The internal simulation time step was taken as \SI{0.1}{ms}.

The synaptic weight between each presynaptic neuron $j$ and the postsynaptic neuron $i$ was initialised randomly at the start of every simulation run, with $w_{ij}$ values uniformly distributed between 0 and $200 / n_i$; as a result, the initial firing rate of the postsynaptic neuron was driven to $\sim \SI{1}{Hz}$.

Input patterns were conveyed to the network by the collective firing activity of presynaptic neurons, where a pattern consisted of a single spike at each neuron. Presynaptic spikes were uniformly distributed over the pattern duration, and selected independently for each neuron. The choice of single rather than multiple input spikes to form pattern representations proved to be more amenable to the subsequent analysis of gathered results. In all cases, an arbitrary realisation of each pattern was used at the start of each simulation run, which was then held fixed thereafter. By this method, a total number $p$ of unique patterns were generated. Patterns were generated with a duration $T = \SI{200}{ms}$, that is approximately the time-scale of sensory processing in the nervous system.

\paragraph{General learning task.}
The postsynaptic neuron was trained to reproduce an arbitrary target output spike train in response to each of the $p$ input patterns through synaptic weight modifications in the network, using either the \ac{INST}, \ac{FILT} or \ac{CHRON} learning rules. In this way, the network learned to perform precise temporal encoding of input patterns. During training, all $p$ input patterns were sequentially presented to the network in batches, where the completion of a batch corresponded to one epoch of learning. Resulting synaptic weight changes computed for each of the individually presented input patterns (or each trial) were accumulated, and applied at the end of an epoch.

The learning rate used for the rules was by default $\eta = 600 / (n_i\; n_s\; p)$, which scaled with the number of presynaptic neurons $n_i$, target output spikes $n_s$ and patterns $p$; any exceptions to this are specified in the main text. As shall be shown in our simulation results, it was indicated that all of the learning rules shared a similar, optimal value for the learning rate.

\paragraph{Performing a single input-output mapping.}
For demonstrative purposes, we first applied the \ac{INST} and \ac{FILT} learning rules to training the network to perform a mapping between a single, fixed input spike pattern and a target output spike train containing four spikes. The network contained 200 presynaptic neurons, and the target output spikes were equally spaced out with timings of \SIlist[list-units = single]{40; 80; 120; 160}{ms}. These wide separations were selected to avoid excessive nonlinear accumulation of error due to interactions between postsynaptic spikes during learning. Simulations for the learning rules were run over 200 epochs, where each epoch corresponded to one, repeated, presentation of the pattern. Hence, a single simulation run reflected \SI{40}{s} of biological time.

\begin{figure}[p]
\includegraphics[max width=\textwidth]{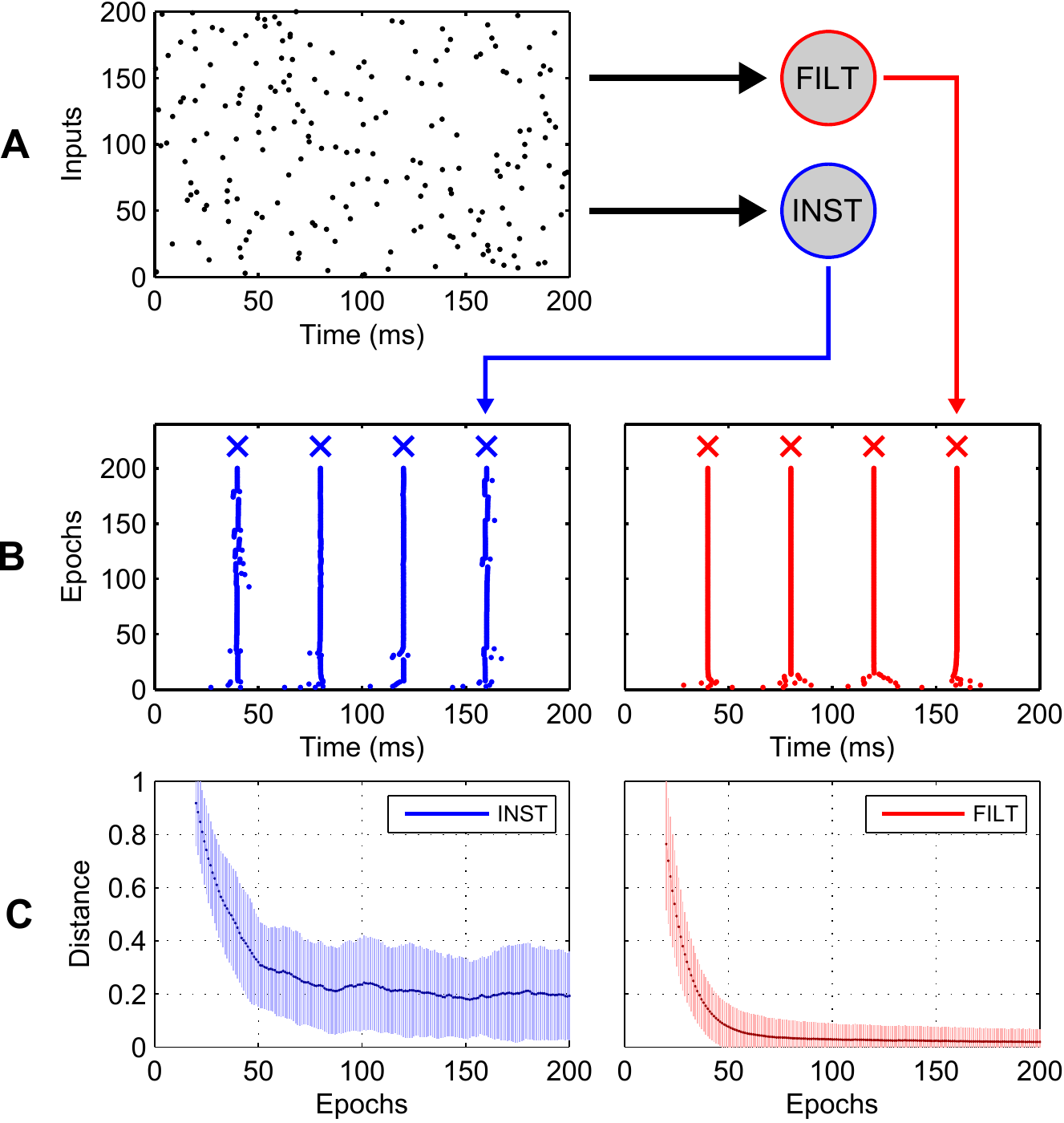}
\caption{
{\bf Two postsynaptic neurons trained under the proposed synaptic plasticity rules, that learned to map between a single, fixed input spike pattern and a four-spike target output train.} (A) A spike raster of an arbitrarily generated input pattern, lasting \SI{200}{ms}, where each dot represents a spike. (B) Actual output spike rasters corresponding to the \ac{INST} rule (left) and the \ac{FILT} rule (right) in response to the repeated presentation of the input pattern. Target output spike times are indicated by crosses. (C) The evolution of the \ac{vRD} for each learning rule, taken as a moving average over 40 independent simulation runs. The shaded regions show the standard deviation.
}
\label{fig5}
\end{figure}

Shown in Fig.~\ref{fig5}A is a spike raster of an arbitrarily generated input pattern, consisting of a single input spike at each presynaptic neuron. In this example, two postsynaptic neurons were tasked with transforming the input pattern into the target output spike train through synaptic weight modifications, as determined by either the \ac{INST} or \ac{FILT} learning rule. From the actual output spike rasters depicted in panel B, it can be seen that both postsynaptic neurons learned to rapidly match their target responses during learning. Despite this, persistent fluctuations in the timings of actual output spikes were associated with just the \ac{INST} rule, while the \ac{FILT} displayed stability over the remaining epochs. Finally, panel C shows the accuracy of each learning rule, given as the average \ac{vRD} (see Eq.~(\ref{eq:vRD})) plotted as a function of the number of learning epochs. With respect to the \ac{INST} rule, it can be seen the \ac{vRD} failed to reach zero and was subject to a high degree of variance, as reflected by the corresponding spike raster in panel B; its final, convergent \ac{vRD} value was \num{0.2 \pm 0.2}, that is an output spike timing error of around \SI{1}{ms} with respect to its target. By contrast, the FILT rule's \ac{vRD} value rapidly approached zero, and was subject to much less variation during the entire course of learning (final \ac{vRD} value was \num{0.02 \pm 0.05}).

\begin{figure}[p]
\includegraphics[max width=\textwidth]{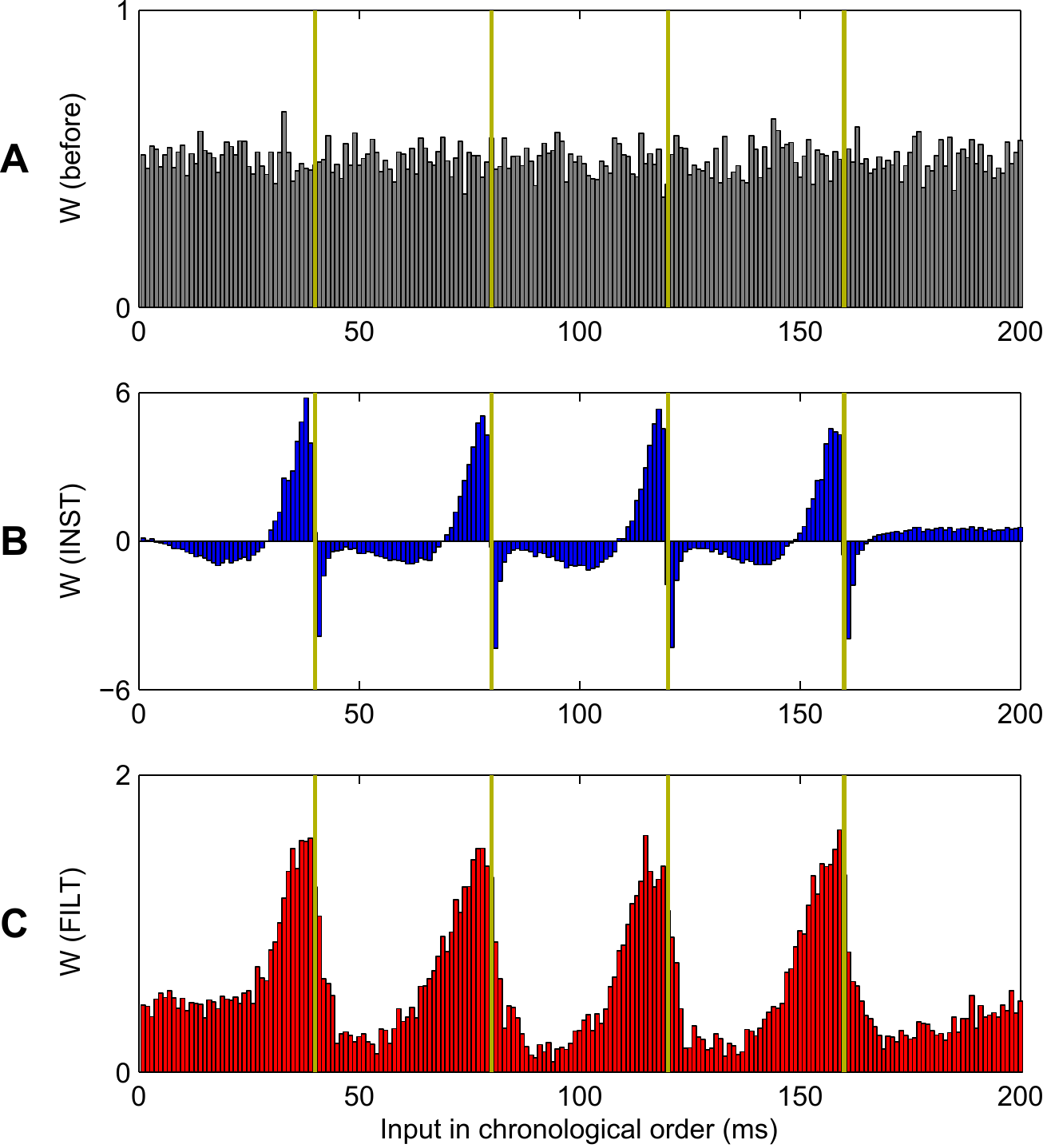}
\caption{
{\bf Averaged synaptic weight values before and after network training, corresponding to the same experiment of Fig.~\ref{fig5}.} The input synaptic weight values are plotted in chronological order, with respect to their associated firing time. (A) The distribution of weights before learning. (B) Post training under the \ac{INST} rule. (C) Post training under the the \ac{FILT} rule. The gold coloured vertical lines indicate the target postsynaptic firing times. Note the different scales of A, B and C. Results were averaged based on 40 independent runs. The design of this figure is inspired from \cite{Mohemmed2012}.
}
\label{fig6}
\end{figure}

\paragraph{Synaptic weight distributions.}
Shown in Fig.~\ref{fig6} are the distributions of synaptic weights before and after network training for the \ac{INST} and \ac{FILT} learning rules, corresponding to the same experiment of Fig.~\ref{fig5}. In plotting Fig.~\ref{fig6}, synaptic weights were sorted in chronological order with respect to their associated presynaptic firing times; for example, the height of a bar at \SI{40}{ms} reflects the average value of a synaptic weight from a presynaptic neuron which transmitted a spike at \SI{40}{ms}. The gold overlaid lines correspond to the previously defined target output spike timings of \SIlist[list-units = single]{40; 80; 120; 160}{ms}.

From this figure, panel A illustrates the uniform distribution of synaptic weights used to initialise the network before any learning took place, which had the effect of driving the initial postsynaptic firing rate to $\sim \SI{1}{Hz}$. Panels B and C show the distribution of synaptic weights at the end of learning, when the \ac{INST} and \ac{FILT} rules were respectively applied. From these two panels, a rapid increase in the synaptic weight values preceding the target output spike timings can be seen, which then proceeded to fall off. Comparatively, the magnitude of weight change was largest for the \ac{INST} rule, with peak values over three times that produced by \ac{FILT}. Furthermore, only the \ac{INST} rule resulted in negatively-valued weights, which is especially noticeable for weights associated with input spikes immediately following the target output spike timings. In effect, these sharp depressions offset the relatively strong input drive received by the postsynaptic neuron just before the target output spike timings, which is indicative of the unstable nature of the \ac{INST} learning rule. By contrast, the \ac{FILT} rule led to a `smoother landscape' of synaptic weight values, following a periodic pattern when plotted in chronological order.

\paragraph{Impact of the learning rate.}
In this experiment we explored the dependence of each rule's performance on the learning rate parameter $\eta$ in terms of the spike-timing accuracy of a trained postsynaptic neuron, measured using the \ac{vRD}. The primary objective was to establish the relative sensitivity of each rule to large values of $\eta$, and secondly to establish a value of $\eta$ which provided a suitable trade-off between learning speed and final convergent accuracy. Here we first include the E-learning \ac{CHRON} rule proposed by \cite{Florian2012}, to provide a benchmark for the \ac{INST} and \ac{FILT} rules. 
With respect to the experimental setup, the network consisted of 200 presynaptic neurons and a single postsynaptic neuron, and was tasked with learning to map a total of 10 different input patterns to the same, single target output spike with a timing of \SI{100}{ms}. In this case learning took place over 500 epochs.

\begin{figure}[t]
\includegraphics[max width=\textwidth]{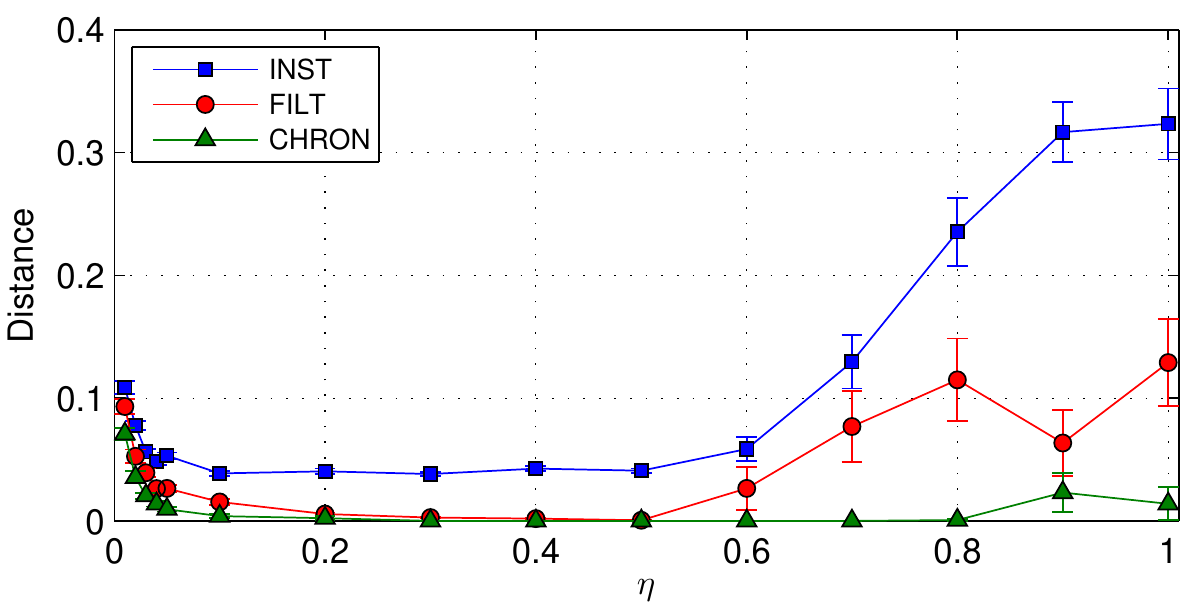}
\caption{
{\bf The \ac{vRD} as a function of the learning rate $\eta$ for each learning rule.} The E-learning \acf{CHRON} rule of \cite{Florian2012} is included as a benchmark for the \ac{INST} and \ac{FILT} rules. In every instance, a network containing 200 presynaptic neurons and a single postsynaptic neuron was tasked with mapping 10 arbitrary input patterns to the same target output spike with a timing of \SI{100}{ms}. Learning took place over 500 epochs, and results were averaged over 40 independent runs. In this case, error bars show the standard error of the mean rather than the standard deviation: the \ac{vRD} was subject to very high variance for large $\eta$ values, therefore we considered just its average value and not its distribution.
}
\label{fig7}
\end{figure}

As shown in Fig.~\ref{fig7} it is clear that the \ac{INST} rule was most sensitive to changes in the learning rate, with an average \ac{vRD} value $2.5 \times$ that of \ac{FILT} for the largest learning rate value $\eta = 1$. The least sensitive rule turned out to be \ac{CHRON}, which still managed to maintain an average \ac{vRD} value close to zero when plotted up to the maximum value of $\eta$. Interestingly, all three distance plots displayed the same general trend over the entire range of learning rates considered: there was a rapid decrease for small $\eta$ values, followed by a plateau up to around $\eta = 0.5$, and then a noticeable increase towards the end. The large distance values for small $\eta$ related to a lack of convergence in learning by the postsynaptic neuron after being trained over 500 epochs.

To summarise, these results support our choice of an identical learning rate for all three learning rules as used in the subsequent learning tasks of this section. Additional, more exhaustive parameter sweeps from further simulations conclusively demonstrated that the learning rates for all three learning rules shared the same inverse proportionality with the number of presynaptic neurons, patterns and target output spikes. This corresponded to an optimal value of $\eta = \num{0.3 \pm 0.1}$ in Fig.~\ref{fig7}.

\paragraph{Classifying spike patterns.}
An important characteristic of a neural network is the maximum number of patterns it can learn to reliably memorise, as well the time taken to train it. Therefore, we tested the performance of the network on a generic classification task, where input patterns belonging to different classes were identified by the precise timings of individual postsynaptic spikes. We first determine the performance of a network when trained to identify separate classes of input patterns based on the precise timing of a \emph{single} postsynaptic spike, and then later consider identifications based on \emph{multiple} postsynaptic spike timings. In this experiment, the network contained a single postsynaptic neuron, and was trained using either the \ac{INST}, \ac{FILT} or \ac{CHRON} learning rule for comparison purposes.

The network was tasked with learning to classify $p$ arbitrarily generated input patterns into five separate classes through hetero-association: an equal number of patterns were randomly assigned to each class, and all patterns belonging to the same class were identified using a shared target output spike timing. Hence, an input pattern was considered to be correctly identified if the postsynaptic neuron responded by firing just a single output spike that fell within $\Delta t$ of its required target timing. The value of $\Delta t$ was varied depending on the level of temporal precision desired, with values selected from the range $\Delta t \in (0, 5]$ ms corresponding to the typical level of spike timing precision as has been observed in the brain \cite{Reich1997}. For each input class a target output spike time was randomly generated according to a uniform distribution that ranged in value between 40 and \SI{200}{ms}; the lower bound of \SI{40}{ms} was enforced, given previous evidence indicating that smaller values are harder to reproduce by an \ac{SNN} \cite{Florian2012,Mohemmed2012}. To ensure input classes were uniquely identified, target output spikes were distanced from each other by a \ac{vRD} of at least 0.5, corresponding to a minimum timing separation of \SI{7}{ms}.

\begin{figure}[p]
\includegraphics[max width=\textwidth]{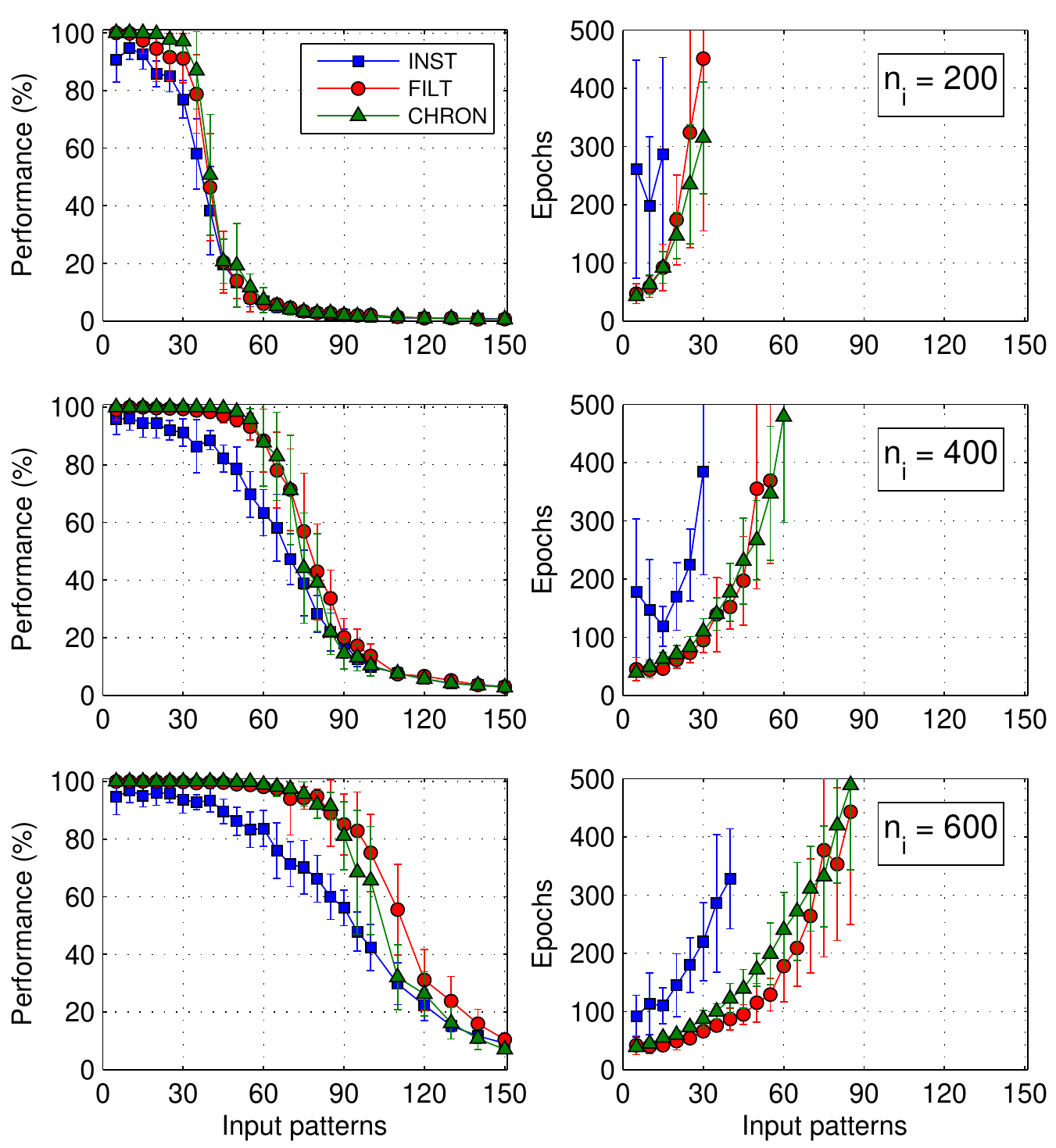}
\caption{
{\bf The classification performance of each learning rule as a function of the number of input patterns when learning to classify $p$ patterns into five separate classes.} Each input class was identified using a single, unique target output spike timing, which a single postsynaptic neuron had to learn to match to within \SI{1}{ms}. {\it Left:} The averaged classification performance $\langle P_c \rangle$ for a network containing $n_i = 200, 400$ and 600 presynaptic neurons. {\it Right:} The corresponding number of epochs taken by the network to reach a performance level of \SI{90}{\%}. More than 500 epochs was considered a failure by the network to learn all the patterns at the required performance level. Results were averaged over 20 independent runs, and error bars show the standard deviation.
}
\label{fig8}
\end{figure}

Shown in the left column of Fig.~\ref{fig8} is the performance of a network containing either 200, 400 or 600 presynaptic neurons, as a function of the number of input patterns to be classified. In this case, we took $\Delta t = \SI{1}{ms}$ as the required timing precision of a postsynaptic spike with respect to its target, for each input class. To quantify the classification performance of the network, we defined a measure $\mathcal{P}_c$ which assumed a value of \SI{100}{\%} in the case of a correct pattern classification, and \SI{0}{\%} otherwise. Hence, in order to determine the maximum number of patterns memorisable by the network, we took an averaged performance level $\langle \mathcal{P}_c \rangle > \SI{90}{\%}$ as our cut-off point when deciding whether all of the patterns were classified with sufficient reliability; this criterion was also used to determine the minimum number of epochs taken by the network to learn all the patterns, and is plotted in the right column of this figure. Epoch values not plotted for an increased number of patterns reflected an inability of the network to learn every pattern within 500 epochs.

As expected, Fig.~\ref{fig8} demonstrates a decrease in the classification performance as the number of input patterns presented to the network was increased, with a clear dependence on the number of presynaptic neurons contained in the network. For example, a network trained under \ac{INST} was able to classify 15, 30 and 40 patterns at a \SI{90}{\%} performance level when containing 200, 400 and 600 presynaptic neurons, respectively. The number of input patterns memorised by a network can be characterised by defining a load factor $\alpha := p / n_i$, where $p$ is the number of patterns presented to a network containing $n_i$ presynaptic neurons \cite{Gutig2006}. Furthermore, the \emph{maximum} number of patterns memorisable by a network can be quantified by its memory capacity $\alpha_m := p_m / n_i$, where $p_m$ is the maximum number of patterns memorised using $n_i$ synapses. Hence, using \SI{90}{\%} as the cut-off point for reliable pattern classifications, we found the \ac{INST} rule had an associated memory capacity of $\alpha_m = \num{0.07 \pm 0.01}$. By comparison, the memory capacities for the \ac{FILT} and \ac{CHRON} rules were $\num{0.14 \pm 0.01}$ and $\num{0.15 \pm 0.01}$, respectively, being around twice the capacity of that determined for \ac{INST}. Beyond these increased memory capacity values, networks trained under \ac{FILT} or \ac{CHRON} were capable of performance levels very close to \SI{100}{\%} when classifying a relatively small number of patterns; by contrast, the maximum performance level attainable under \ac{INST} was just over \SI{95}{\%}, and was subject to a relatively large variance of around \SI{5}{\%}. Finally, it is evident from this figure that both \ac{FILT} and \ac{CHRON} shared roughly the same performance levels over the entire range of input patterns and network structures considered. In terms of the time taken to train the network, both \ac{FILT} and \ac{CHRON} were equally fast, while \ac{INST} was typically slower than the other rules by a factor of between three and four. This difference in the training time became more pronounced as both the number of input patterns and presynaptic neurons were increased.

\paragraph{Memory capacity.}
We now explore in more detail the memory capacity $\alpha_m$ supported under each learning rule, specifically with respect to its dependence on the output spike timing precision $\Delta t$ used to identify input patterns. In determining the memory capacity as a function of the timing precision, we used the same experimental setup as considered previously for $\Delta t = \SI{1}{ms}$, but extended to also consider values of $\Delta t$ between 0.2 and \SI{5}{ms} (equally spaced in increments of \SI{0.2}{ms}). As before, we assumed the maximum number of patterns memorisable by the network as those that were classified with a corresponding averaged classification performance $\langle \mathcal{P}_c \rangle$ of at least \SI{90}{\%} within 500 epochs.

\begin{figure}[t]
\includegraphics[max width=\textwidth]{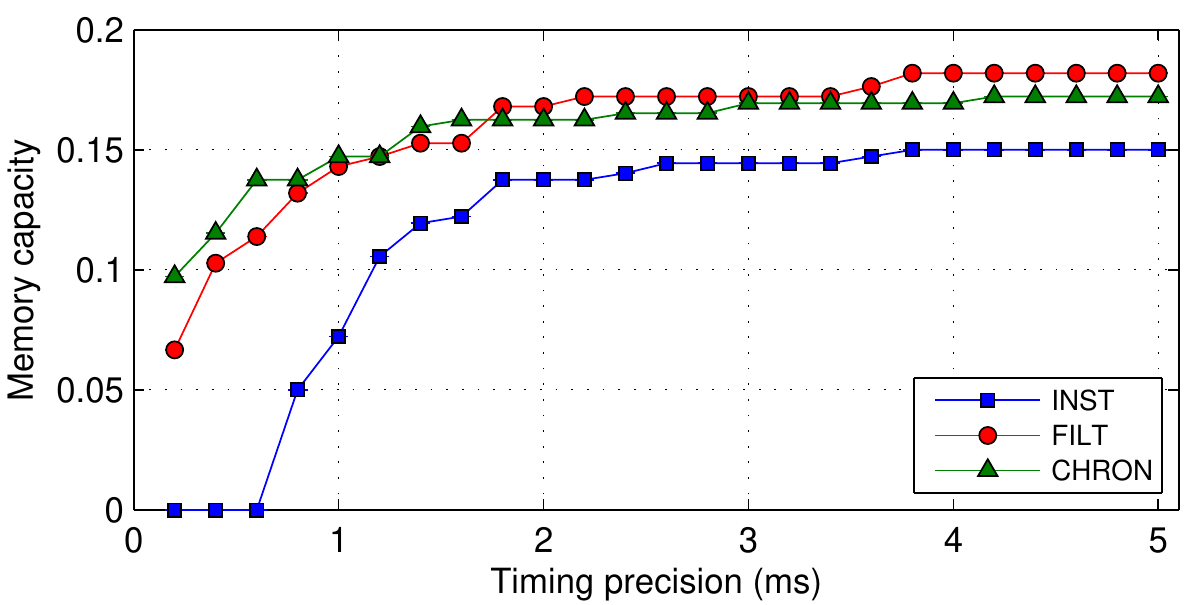}
\caption{
{\bf The memory capacity $\alpha_m$ of each learning rule as a function of the required output spike timing precision.} The network contained a single postsynaptic neuron, and was trained to classify input patterns into five separate classes within 500 epochs. Memory capacity values were determined based on networks containing $n_i = 200$, 400 and 600 presynaptic neurons. Results were averaged over 20 independent runs.
}
\label{fig9}
\end{figure}

From Fig.~\ref{fig9} it can be seen that the memory capacity provided by each learning rule increased with the size of the timing precision, which eventually levelled off for values $\Delta t > \SI{3}{ms}$. It is also clear that the trend for the \ac{FILT} rule is consistent with that for \ac{CHRON} over the entire range of timing precision values considered, while the \ac{INST} rule gave rise to the lowest memory capacities. For values $\Delta t < \SI{2}{ms}$ the difference in memory capacity between \ac{INST} and \ac{FILT} was most pronounced, to the extent that \ac{INST} was incapable of memorising any input patterns for $\Delta t < \SI{0.8}{ms}$. By contrast, \ac{FILT} still maintained a memory capacity close to 0.07 when classifying patterns based on ultra-precise output spike timings of within \SI{0.2}{ms}. As a validation of our method, we note that our measured memory capacity for \ac{CHRON} at a timing precision of \SI{1}{ms} is in close agreement with that determined originally in Fig.~9A of \cite{Florian2012}: with a value close to 0.15 after 500 epochs of network training.

\paragraph{Multiple target output spikes.}
Finally, we examine the performance of the learning rules when input patterns are identified by the timings of \emph{multiple} postsynaptic spikes. In this case, the network contained 200 presynaptic neurons and a single postsynaptic neuron, and was trained to classify a total of 10 input patterns into five separate classes, with two patterns belonging to each class. Both patterns belonging to a class were identified by the same target output spike train; hence, a correct pattern classification was considered when the number of actual output spikes fired by the postsynaptic neuron matched the number of target output spikes, and every actual spike fell within $\Delta t$ of its respective target. For each input class, target output spikes were randomly generated according to a uniform distribution bound between 40 and \SI{200}{ms}, as used previously. A minimum inter-spike interval of \SI{10}{ms} was enforced to minimise interactions between output spikes. To ensure input classes were uniquely represented, generated target output spike trains were distanced from one another by a \ac{vRD} of at least $n_s / 2$, where $n_s$ was the number of spikes contained in a target train.

\begin{figure}[t]
\includegraphics[max width=\textwidth]{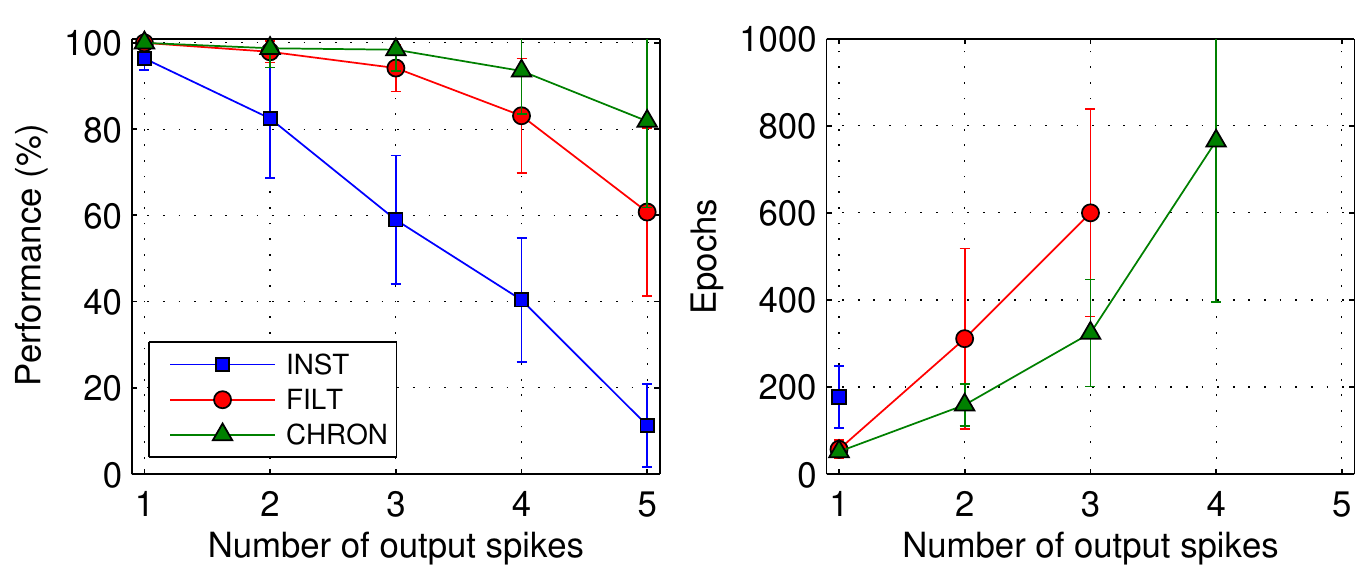}
\caption{
{\bf The classification performance of each learning rule as a function of the number of target output spikes used to identify input patterns.} The network was tasked when classifying 10 input patterns into 5 separate classes. Correct classifications were considered when the number of actual output spikes fired by a single postsynaptic neuron matched that of its target, and each actual spike fell within \SI{1}{ms} of its corresponding target timing. In this case, a network containing 200 presynaptic neurons was trained over an extended 1000 epochs to allow for decreased learning speed, and results were averaged over 20 independent runs.
}
\label{fig10}
\end{figure}

Shown in Fig.~\ref{fig10} is the performance of the network trained under each learning rule when classifying input patterns based on the precise timings of between one and five target output spikes, with a timing precision $\Delta t = \SI{1}{ms}$. Because the learning rate was inversely proportional to the number of target spikes, we extended the maximum number of epochs to 1000 to ensure the convergence of each rule. As can be seen in this figure, the performance dropped as the number of output spikes increased, and most noticeably for the \ac{INST} rule which returned a minimum performance value approaching \SI{0}{\%} when patterns were identified using five output spikes. By comparison, the \ac{CHRON} rule gave rise to the highest performance levels over the entire range of output spikes tested, closely followed by the \ac{FILT} rule. If we count the maximum number of output spikes learnable by the network above a \SI{90}{\%} performance level, we obtain one, three and four output spikes for \ac{INST}, \ac{FILT} and \ac{CHRON}, respectively, where the associated number of training epochs in each instance is plotted in the right panel of the figure. From this, it is observed that \ac{CHRON} was fastest in training the network to learn multi-spike based pattern classifications, closely followed by \ac{FILT} and finally \ac{INST}.

\paragraph{Summary.}
Taken together, the experimental results of this section demonstrate a similarity in the performance between the \ac{FILT} and \ac{CHRON} rules under most circumstances, except when applied to learning multiple target output spikes for which the \ac{CHRON} rule was best suited. The \ac{INST} rule, however, performed worst in all cases, and in particular displayed difficulties when classifying input patterns with increasingly fine temporal precision. This disparity between \ac{INST} and the other two rules is explained by its unstable behaviour, since it essentially fails to account for the temporal proximity of neighbouring target and actual postsynaptic spikes. As was predicted in our earlier analysis, this instability gave rise to fluctuating postsynaptic spikes close to their target timings (see Fig.~\ref{fig5}). Hence, it is evident that exponentially filtering postsynaptic spikes in order to drive more gradual synaptic weight modifications confers a strong advantage when temporally precise encoding of input patterns is desired.

From the experiment concerning pattern classifications based on multiple output spike timings, it was found for each of the learning rules that the performance decreased with the number of target output spikes. This is not surprising given that the network needed to match every one of its targets with the same level of temporal precision, effectively increasing the synaptic load of the network during learning. Qualitatively, these results are consistent with those found in \cite{Florian2012} for the E-learning \ac{CHRON} rule.

\section*{Discussion}

We have studied the conditions under which supervised synaptic plasticity can most effectively be applied to training \acp{SNN} to learn precise temporal encoding of input patterns. For this purpose, we have derived two supervised learning rules, termed \ac{INST} and \ac{FILT}, and analysed the validity of their solutions on several, generic, input-output spike timing association tasks. We have also tested the proposed rules' performance in terms of the maximum number of spatio-temporal input patterns that a trained network can memorise per synapse, with patterns identified based on the precise timing of an output spike emitted by a postsynaptic neuron; this experiment was designed to reflect the experimental observations of biological neurons which utilise the relative timing of their output spike for stimulus encoding \cite{Gollisch2008,Johansson2004}. In order to benchmark the performance of our proposed rules, we also implemented the previously established E-learning \ac{CHRON} rule.
From our simulations, we found \ac{FILT} approached the high performance level of \ac{CHRON}: relating to its ability to smoothly converge towards stable, desired solutions by account of its exponential filtering of postsynaptic spike trains. By contrast, \ac{INST} consistently returned the lowest performance, which was underpinned by its tendency to result in fluctuations of emitted postsynaptic spikes near their target timings.

Essentially, weight changes driven by the \ac{INST} and \ac{FILT} rules depend on a combination of two activity variables: a postsynaptic error term to signal appropriate output responses, and a presynaptic eligibility term to capture the coincidence of input spikes with the output error. \ac{INST} and \ac{FILT} differ, however, with respect to their postsynaptic error term: while \ac{INST} relies on the instantaneous difference between target and actual output spike trains, \ac{FILT} instead relies on the smoothed difference between exponentially filtered target and actual output spike trains. Despite this, both rules share the same presynaptic eligibility term, that is the \ac{PSP} evoked due to an input spike. In our analysis, the \ac{PSP} was determined as a suitable presynaptic factor, whereas the structurally similar \ac{SPAN} and \ac{PSD} rules instead rely on an arbitrarily defined presynaptic kernel that is typically related to the neuron's postsynaptic current \cite{Mohemmed2012,Yu2013}. Interestingly, in the authors' analysis of the \ac{SPAN} rule an $\alpha$-shaped kernel was indicated as providing the best performance during learning, which closely resembles the shape of a \ac{PSP} curve as used here.

In our analysis of single synapse dynamics (see Results section), we predicted the \ac{FILT} rule to provide convergence towards a stable and desired synaptic weight solution, offering an explanation for its high performance as tested through subsequent simulations in large network sizes. The key advantage of the \ac{FILT} rule is its ability to signal not just the timings of desired or erroneous postsynaptic spikes, but also their temporal proximity with other postsynaptic spikes as measured via their convolution with an exponential kernel; in this way, the \ac{FILT} rule is able to smoothly align emitted postsynaptic spikes with their respective targets by avoiding unstable synaptic weight changes. This operation is roughly analogous to one used by the E-learning \ac{CHRON} rule, which includes a distinct mechanism for carefully shifting actual postsynaptic spikes towards their neighbouring targets, making it a highly efficient spike-based neural classifier \cite{Florian2012}.
The \ac{FILT} and \ac{CHRON} rules differ, however, in terms of their implementation: while \ac{FILT} can potentially be implemented as an online-based learning method for biological realism, \ac{CHRON} is restricted to offline learning, given that it depends on discrete summations over cost functions that are non-local in time as derived from the \ac{VPD} measure. Comparatively, the \ac{INST} rule was predicted to provide imperfect and unstable convergence during learning, which we attributed to its inability to effectively account for neighbouring target and actual postsynaptic spikes.

Computer simulations were run to test the performance of the \ac{INST} and \ac{FILT} rules in terms of their temporal encoding precision in large network sizes, including the E-learning \ac{CHRON} rule for comparison purposes. We found \ac{FILT} and \ac{CHRON} were consistent with each other performance-wise, and largely outperformed \ac{INST}. It is worth pointing out, however, that \ac{FILT} is more straightforward to implement than \ac{CHRON}, since it avoids the added complexity of having to establish whether target and actual postsynaptic spikes are independent of each other or not based on the \ac{VPD} measure \cite{Florian2012}. By comparison, \ac{INST} is the simplest rule to implement, but comes at the cost of significantly decreased spike timing precision.

On all these learning tasks neurons were trained to classify input patterns using the precise timings of output spikes; an alternative and more practical method for classifying patterns might instead take the minimum distance between target and actual output spike trains in order to discriminate between different input classes, which would more effectively counteract misclassifications in the case of input noise \cite{Gardner2015}. In this work, however, we adopted a classification method based on the precise timings of output spikes for the sake of consistency with more directly related previous studies \cite{Florian2012,Mohemmed2012,Yu2013}, and to more thoroughly compare the relative performance of each learning rule with respect to the precision of their temporal encoding.

\subsection*{Related Work}

In our approach, we started by taking gradient ascent on an objective function for maximising the likelihood of generating desired output spike trains, based on the statistical method of \cite{Pfister2006}; this method is well suited to our analysis, especially since it has been shown to have a unique global maximum that is obtainable using a standard gradient ascent procedure \cite{Paninski2004}. Next, we substituted the stochastic spiking neuron model used during the derivation with a deterministic \ac{LIF} neuron model, such that output spikes were instead restricted to being generated upon crossing a fixed firing threshold. In this way, the resulting \ac{INST} and \ac{FILT} rules have a reasonably strong theoretical basis, rely on intrinsic neuronal dynamics, and further still allow for the efficient learning of desired sequences of precisely-timed output spikes. By comparison, most previous approaches to formulating supervised learning rules for \acp{SNN} have relied on heuristic approximations, such as adapting the Widrow-Hoff rule for use with spiking neuron models \cite{Ponulak2010,Mohemmed2012,Yu2013}, or mapping from Perceptron to spike-based learning \cite{Albers2013,Xu2013}. Moreover, although the well known \acf{ReSuMe} \cite{Ponulak2010} can more rigorously be reinterpreted as a gradient descent learning procedure \cite{Sporea2013}, assumptions are still made regarding the functional dependence of weight changes on the relative timing differences between spikes, for the purposes of mimicking a Hebbian-like \ac{STDP} rule \cite{Gerstner2002}.

According to the study of \cite{Memmesheimer2014}, the upper limit on the number of input-output pattern transformations a spiking neuron can learn to memorise falls between 0.1 and 0.3 per synapse, based on single target output spikes. In establishing this maximal capacity estimate, the authors of this study applied an idealised \acf{HTP} learning method, such that the firing times of the trained neuron were enforced at its target timings. From Fig.~\ref{fig9}, we determined the \ac{FILT} and \ac{CHRON} rules to approximately share the same memory capacity, with measured values between 0.15 and 0.2 for required timing precisions larger than \SI{1}{ms}; hence, the capacities afforded by these two rules can be regarded as approaching maximal values. By contrast, the \ac{INST} rule only remained competitive with \ac{FILT} and \ac{CHRON} for relatively large values of the required timing precision, with values of at least $\sim \SI{3}{ms}$. It is noted that our capacity measurements here do not reflect upper estimates; in our approach, networks were trained over a maximum of 500 epochs to also test the rapidity of learning, whereas previous studies have trained networks, for instance using E-learning \ac{CHRON}, over up to two orders of magnitude increased duration \cite{Florian2012,Albers2016}.

The authors of \cite{Memmesheimer2014} also presented an \ac{FP} learning rule that more realistically depends on intrinsic neuronal dynamics, unlike their \ac{HTP} method. As discussed previously, the \ac{FP} rule is essentially an \ac{INST}-like method, in the sense that weight updates depend on an instantaneously communicated error signal. However, \ac{FP} differs by just taking into account the first error during learning in each trial. The \ac{FP} rule has been shown to provide a high memory capacity that is comparable with \ac{HTP}, as well as having been proven to converge towards a stable solution in finite time. By comparison, our results have demonstrated reduced performance for \ac{INST}, and in particular for small values of the required timing precision, $\Delta t$. Despite this, our \ac{INST} has the potential for being implemented as a fully online method in simulations, unlike \ac{FP} which must be immediately shut down upon encountering the first error during a trial. It would be interesting to explore an online implementation of \ac{FP} learning for increased biological plausibility, while maintaining its high performance by minimising nonlinear interactions between output error signals. Realistically, this might be realised by introducing a refractory effect in the neuron's error signals \cite{Memmesheimer2014}.

It is highlighted that the \ac{INST} and \ac{FILT} rules are capable of learning \emph{multiple} target output spikes; this is an important feature of any spike-based learning rule, and makes them more biologically relevant considering that precise spike timings represent a more fundamental unit of computation in the nervous system than that of lengthier firing rates \cite{VanRullen2005}. Multi-spike learning rules are a natural progression from single-spike rules, such as from the original SpikeProp algorithm which is restricted to learning single-spike target outputs \cite{Bohte2002}, and the Tempotron which is only capable of learning to either fire or not-fire an output spike \cite{Gutig2006}.

\subsection*{Biological Plausibility}

Out of the rules studied here, we believe \ac{FILT} matches most criteria to be considered of biological relevance: first, weight updates depend on pre- and postsynaptic activity variables that are locally available at each synapse. Second, its postsynaptic error term is communicated by a smoothly decaying signal that is based on the difference between filtered target and actual output spikes, which might arise from the concentration of a synaptic neuromodulator influenced by backpropagated action potentials \cite{Bush2012}. Finally, it is implementable as an online learning method, which is important when considering how information is most likely processed continuously by the nervous system.

As with most existing learning rules for \acp{SNN}, the proposed rules depend on the presence of a supervisory signal to guide synaptic weight modifications. A possible explanation for supervised learning might come from so termed `referent activity templates', or spike patterns generated by neural circuits existing elsewhere in the brain, which are to be mimicked by circuits of interest during learning \cite{Knudsen1994,Miall1996}. A detailed model of supervised learning in \acp{SNN} has recently been proposed by \cite{Urbanczik2014}, providing a strong mechanistic explanation for how such referent activity templates might be used to drive the learning of desired postsynaptic activity patterns. Specifically, this method has utilised a compartmental model, simulating the somatic and dendritic dynamics of a stochastic spiking neuron, such that the neuron's firing activity is determined by integrating its direct input from somatic synapses with its network input via dendritic synapses. In this way, the neuron's firing activity can be directly `nudged' towards some desired pattern via its somatic input (or template pattern), while plasticity at its dendritic synapses takes care of forming associations of this target activity pattern with input patterns that are simultaneously presented to the network. 
The \ac{INST} and \ac{FILT} synaptic plasticity rules here can in principle be implemented based on this compartmental model for increased biological realism.

A more recent study \cite{Albers2016} has also drawn inspiration from such an associative learning paradigm, culminating in a synaptic plasticity rule that works to maintain a neuron's membrane potential below its firing threshold during learning, termed \ac{MPDP}. Essentially, \ac{MPDP} is an unsupervised learning rule, and by itself is used to train a neuron to remain quiescent in response to input activity. However, if \ac{MPDP} is also combined with strong synaptic input, delivered from an external source, that is briefly injected into a trained neuron at its desired firing timings, then the rule instead functions as a supervised one. Similarly to the study by \cite{Urbanczik2014}, \ac{MPDP} demonstrates how the supervised learning of precisely timed spikes in an \ac{SNN} can arise in a biologically meaningful way.

A final possibility, and one that is gaining increasing interest, is that supervised signalling might actually reflect a form of reinforcement-based learning, but operating on a shorter time-scale. Several, biologically meaningful learning rules have been proposed based on reward-modulated synaptic plasticity \cite{Izhikevich2007,Farries2007,Fremaux2010}, including a reimplementation of Elman backpropagation \cite{Gruning2007}, and in our previous work we have successfully demonstrated how reinforcement learning can be applied to learning multiple, and precisely timed, output spikes \cite{Gardner2013}.

\section*{Conclusions}

In this paper, we have addressed the scarcity of existing learning rules for networks of spiking neurons that have a theoretical basis, and which allow for the learning of \emph{multiple} and \emph{precisely-timed} output spikes. In particular, we have shown our proposed \ac{FILT} rule, which is based on exponentially filtered output spike trains, to be a highly efficient, spike-based neural classifier. Classifiers based on a temporal code are of interest since they are theoretically more capable than those using a rate-based code when processing information on rapid time-scales.

In our analysis, we have restricted our attention to relatively small network sizes when testing the performance of the proposed learning rules. Our main intention, though, was to explore their potential for driving accurate synaptic weight modifications, rather than the scaling of their performance with an increasing number of input synapses. However, it would be of increased biological significance to test the performance of a learning method as applied to a much larger network size: containing on the order of $10^4$ synapses per neuron as is typical in the nervous system. Practically, this could well be achieved via implementation in neuromorphic hardware, such as the massively-parallel computing architecture of SpiNNaker \cite{Furber2014}. As a starting point, the simplistic \ac{INST} rule could be implemented in SpiNNaker, representing an achievable, and exciting, aim for future work.

\section*{Acknowledgments}

This work was supported by the Engineering and Physical Sciences Research Council (EPSRC, grant no. EP/J500562/1), the European Community's Seventh Framework Programme (FP7/2007-2013, grant no. 604102, HBP -- the Human Brain Project) and Horizon 2020 (grant no. 284941, HBP).

\section*{Appendix}\label{app}

\setcounter{equation}{0}
\renewcommand{\theequation}{A\arabic{equation}}

\paragraph{Convergence of gradient ascent procedure based on intrinsic neuronal dynamics.}
In our approach we wish to consider a learning rule that depends on the intrinsic dynamics of a postsynaptic neuron, rather than artificially clamping its firing activity to its target response as in Eq.~(\ref{eq:w_update_stoch}). This formula is restated below: 
\begin{equation} \label{eq:w_update_intrinsic_supp}
\Delta w_{ij} = \frac{\eta}{\Delta u} \int_0^T \left[ \mathcal{Y}_i^{\mathrm{ref}}(t) - \rho_i(t|\mathbf{x}, y_i) \right] \sum_{t_j^f \in x_j} \epsilon (t - t_j^f)\, \mathrm{d}t \;,
\end{equation}
where we have substituted $\rho_i(t|\mathbf{x}, y_i^\mathrm{ref})$ with $\rho_i(t|\mathbf{x}, y_i)$, such that the stochastic intensity of the postsynaptic neuron depends on its actual sequence of emitted output spikes $y_i$ rather than its target output $y_i^\mathrm{ref}$. We show here that Eq.~(\ref{eq:w_update_intrinsic_supp}) yields similar weight updates to Eq.~(\ref{eq:w_update_stoch}) if the actual postsynaptic spike train is already close to its target. 

To demonstrate this, we start by considering the absolute difference between weight updates applied using Eqs.~(\ref{eq:w_update_intrinsic_supp}) and (\ref{eq:w_update_stoch}):
\begin{equation} \label{eq:converge_1}
\left| \Delta w_{ij} - \Delta w_{ij}^\mathrm{ref} \right| = \frac{\eta}{\Delta u} \bigg| \int_0^T \left[ \rho_i(t|\mathbf{x}, y_i) - \rho_i(t|\mathbf{x}, y_i^\mathrm{ref}) \right] \sum_{t_j^f \in x_j} \epsilon (t - t_j^f)\, \mathrm{d}t \bigg| \;,
\end{equation}
leading to the following inequality for an absolute integrand:
\begin{equation} \label{eq:converge_2}
|\Delta w_{ij} - \Delta w_{ij}^\mathrm{ref}| \leq \frac{\eta}{\Delta u} \int_0^T \left| \rho_i(t|\mathbf{x}, y_i) - \rho_i(t|\mathbf{x}, y_i^\mathrm{ref}) \right| \sum_{t_j^f \in x_j} \epsilon (t - t_j^f)\, \mathrm{d}t \;.
\end{equation}
Now, for simplicity, if we assume one of the presynaptic neurons, denoted $j$, contributes a single input spike at time $t_j = \SI{0}{ms}$, and a single target and actual output spike occur at times $t_i^\mathrm{ref}$ and $t_i$ for a postsynaptic neuron $i$, respectively, then the above equation simplifies to
\begin{equation} \label{eq:converge_3}
|\Delta w_{ij} - \Delta w_{ij}^\mathrm{ref}| \leq \frac{\eta}{\Delta u} \int_0^T \left| \rho_i(t|\mathbf{x}, t_i) - \rho_i(t|\mathbf{x}, t_i^\mathrm{ref}) \right| \epsilon (t)\, \mathrm{d}t \;.
\end{equation}
Here, $\rho_i(t|\mathbf{x}, t_i)$ denotes a dependence of the postsynaptic neuron's stochastic intensity at time $t$ on the entire set of presynaptic spikes $\mathbf{x}$, including from neuron $j$, and its actual output firing time $t_i$.

According to the definition of the \ac{PSP} kernel in Eq.~(\ref{eq:PSP_kernel}), the kernel assumes a maximum value, denoted $\epsilon^\mathrm{peak}$. Hence, an upper bound of Eq.~(\ref{eq:converge_3}) can be given by
\begin{equation} \label{eq:converge_4}
|\Delta w_{ij} - \Delta w_{ij}^\mathrm{ref}| \leq \frac{\eta\, \epsilon^\mathrm{peak}}{\Delta u} \int_0^{T} \left| \rho_i(t|\mathbf{x}, t_i) - \rho_i(t|\mathbf{x}, t_i^\mathrm{ref}) \right|\, \mathrm{d}t \;.
\end{equation}
We emphasise here that although we consider just a single input spike, the above would equally be valid for multiple input spikes by simply multiplying the upper bound on the \ac{PSP}, $\epsilon^\mathrm{peak}$, by the number of spikes contributed from neuron $j$.

As defined by Eq.~(\ref{eq:EXP_rate}) the stochastic intensity has an exponential dependence on the postsynaptic neuron's membrane potential, and the only difference between $\rho_i$ and $\rho_i^\mathrm{ref}$ arises from their reset term $\kappa$, hence Eq.~(\ref{eq:converge_4}) can be written as
\begin{equation} \label{eq:converge_5}
|\Delta w_{ij} - \Delta w_{ij}^\mathrm{ref}| \leq \frac{\eta\, \epsilon^\mathrm{peak}}{\Delta u} \int_0^{T} \rho_i(t|\mathbf{x}, t_i^\mathrm{ref})\, \left| \exp\left( \frac{ \kappa(t - t_i) - \kappa(t - t_i^\mathrm{ref}) }{ \Delta u } \right) - 1  \right|\, \mathrm{d}t \;.
\end{equation}
For a given finite set of inputs (all predecessor neurons of $i$) on the finite interval $[0,T]$, $u_i$ is smaller than a constant, irrespective of where the target and actual output spikes fall (consider the maximum of $u_i$ in Eq.~\eqref{eq:potential} for $y_i$ the empty set). Therefore, $\rho_i$ is also bounded by a constant $\rho_i^{\mathrm{max}}$, and hence
\begin{equation} \label{eq:converge_6}
|\Delta w_{ij} - \Delta w_{ij}^\mathrm{ref}| \leq \frac{\eta\, \epsilon^\mathrm{peak}\, \rho_i^\mathrm{max}}{\Delta u} \int_0^{T} \left| \exp\left( \frac{ \kappa(t - t_i) - \kappa(t - t_i^\mathrm{ref}) }{ \Delta u } \right) - 1 \right|\, \mathrm{d}t \;.
\end{equation}
We now show in the following that the integral on the right-hand side of the above equation, and by extension the difference in the weight change, becomes small if $t_i$ and $t_i^\mathrm{ref}$ are close together. If we assume $\delta t := t_i - t_i^\mathrm{ref} > 0$ (an analogous argument applies also for $\delta t < 0$), then the difference between the reset kernels can be expressed as
\begin{align} \label{eq:converge_7}
K(t) &:= \kappa(t - t_i) - \kappa(t- t_i^{\mathrm{ref}}) \;, \\
\intertext{and the absolute difference}
|K(t)| &= | \kappa_0| \left| \exp \left( -\frac{t-t_i}{\tau_m} \right) \Theta(t - t_i) - \exp\left( -\frac{t-t_i^\mathrm{ref}}{\tau_m} \right) \Theta(t - t_i^\mathrm{ref}) \right| \nonumber \\
&\leq | \kappa_0| \left| \exp \left( -\frac{t-t_i}{\tau_m} \right) - \exp\left( -\frac{t-t_i^\mathrm{ref}}{\tau_m} \right) \right|  \Theta(t - t_i) \nonumber \\
&\quad + | \kappa_0 | \exp\left( -\frac{t-t_i^\mathrm{ref}}{\tau_m} \right) \mathbbm{1}_A(t) \;,
\end{align}
where $\mathbbm{1}_A(t)$ is the indicator function on the interval $A = [t_i^\mathrm{ref},t_i]$, such that $\mathbbm{1}_A(t) = 1$ if $t \in A$, and $\mathbbm{1}_A(t) = 0$ otherwise. Furthermore, the following inequality applies:
\begin{equation} \label{eq:converge_8}
|K(t)| \leq |\kappa_0| \exp\left( -\frac{t-t_i}{\tau_m} \right) \left| 1 - \exp\left( -\frac{\delta t}{\tau_m}\right) \right| \Theta(t - t_i)  + | \kappa_0| \mathbbm{1}_A(t) \;.
\end{equation}
Hence, $|K(t)|$ tends to zero point-wise in $t$ for $\delta t \to 0$ and is bounded by $|\kappa_0|$. By continuity, the integrand in Eq.~\eqref{eq:converge_6}, $|\exp(K(t)) - 1|$, also goes to zero pointwise and is bounded. Hence by dominated convergence, the integral in \eqref{eq:converge_6}, tends to zero for $\delta t \rightarrow 0$. In other words, it is continuous in $\delta t$, and we can find for each $\varepsilon$ a $\delta t$ such that weight changes based on $t_i$ do not differ by more than $\varepsilon$ from those based on $t_i^\mathrm{ref}$ if target and actual output spikes are closer than $\delta t$.

The proof above shows that the intrinsic weight update rule of Eq.~\eqref{eq:w_update_intrinsic} yields comparable weight changes to the rule in Eq.~\eqref{eq:w_update_stoch} if the actual output spike is already close to its target. However, in the form given above it is not constructive, i.e. does not give us an explicit estimate of $\delta t$ in terms of $\varepsilon$ and other parameters. As for all learning rules, their practical feasibility has to be demonstrated in simulations.


\begin{thebibliography}{10}

\bibitem{VanRullen2005}
van Rullen R, Guyonneau R, Thorpe SJ.
\newblock Spike times make sense.
\newblock Trends in Neurosciences. 2005;28(1):1--4.

\bibitem{Gollisch2008}
Gollisch T, Meister M.
\newblock Rapid neural coding in the retina with relative spike latencies.
\newblock Science. 2008;319(5866):1108--1111.

\bibitem{Johansson2004}
Johansson RS, Birznieks I.
\newblock First spikes in ensembles of human tactile afferents code complex
  spatial fingertip events.
\newblock Nature Neuroscience. 2004;7(2):170--177.

\bibitem{Mainen1995}
Mainen ZF, Sejnowski TJ.
\newblock Reliability of spike timing in neocortical neurons.
\newblock Science. 1995;268(5216):1503--1506.

\bibitem{Reich1997}
Reich DS, Victor JD, Knight BW, Ozaki T, Kaplan E.
\newblock Response variability and timing precision of neuronal spike trains in
  vivo.
\newblock Journal of Neurophysiology. 1997;77(5):2836--2841.

\bibitem{Uzzell2004}
Uzzell V, Chichilnisky E.
\newblock Precision of spike trains in primate retinal ganglion cells.
\newblock Journal of Neurophysiology. 2004;92(2):780--789.

\bibitem{Kasinski2006}
Kasinski A, Ponulak F.
\newblock Comparison of supervised learning methods for spike time coding in
  spiking neural networks.
\newblock Int J Appl Math Comput Sci. 2006;16(1):101--113.

\bibitem{Gutig2014}
G{\"u}tig R.
\newblock To spike, or when to spike?
\newblock Current Opinion in Neurobiology. 2014;25:134--139.

\bibitem{Mohemmed2012}
Mohemmed A, Schliebs S, Matsuda S, Kasabov N.
\newblock {SPAN}: Spike pattern association neuron for learning spatio-temporal
  spike patterns.
\newblock International Journal of Neural Systems. 2012;22(04).

\bibitem{Yu2013}
Yu Q, Tang H, Tan KC, Li H.
\newblock Precise-spike-driven synaptic plasticity: Learning hetero-association
  of spatiotemporal spike patterns.
\newblock PLoS ONE. 2013;8(11):e78318.

\bibitem{Florian2012}
Florian RV.
\newblock The Chronotron: A Neuron That Learns to Fire Temporally Precise Spike
  Patterns.
\newblock PLoS ONE. 2012;7(8):e40233.

\bibitem{Victor1996}
Victor JD, Purpura KP.
\newblock Nature and precision of temporal coding in visual cortex: a
  metric-space analysis.
\newblock Journal of Neurophysiology. 1996;76(2):1310--1326.

\bibitem{Memmesheimer2014}
Memmesheimer RM, Rubin R, {\"O}lveczky BP, Sompolinsky H.
\newblock Learning precisely timed spikes.
\newblock Neuron. 2014;82(4):925--938.

\bibitem{Pfister2006}
Pfister JP, Toyoizumi T, Barber D, Gerstner W.
\newblock Optimal spike-timing-dependent plasticity for precise action
  potential firing in supervised learning.
\newblock Neural Computation. 2006;18(6):1318--1348.

\bibitem{Bi1998}
Bi G, Poo M.
\newblock Synaptic modifications in cultured hippocampal neurons: dependence on
  spike timing, synaptic strength, and postsynaptic cell type.
\newblock The Journal of Neuroscience. 1998;18(24):10464--10472.

\bibitem{Fremaux2010}
Fr{\'e}maux N, Sprekeler H, Gerstner W.
\newblock Functional requirements for reward-modulated spike-timing-dependent
  plasticity.
\newblock The Journal of Neuroscience. 2010;30(40):13326--13337.

\bibitem{Gardner2015}
Gardner B, Sporea I, Gr{\"u}ning A.
\newblock Learning Spatiotemporally Encoded Pattern Transformations in
  Structured Spiking Neural Networks.
\newblock Neural Computation. 2015;27(12):2548--2586.

\bibitem{Brea2013}
Brea J, Senn W, Pfister JP.
\newblock Matching recall and storage in sequence learning with spiking neural
  networks.
\newblock The Journal of Neuroscience. 2013;33(23):9565--9575.

\bibitem{JimenezRezende2014}
Rezende DJ, Gerstner W.
\newblock Stochastic variational learning in recurrent spiking networks.
\newblock Frontiers in Computational Neuroscience. 2014;8.

\bibitem{Gardner2013}
Gardner B, Gr{\"u}ning A.
\newblock Learning Temporally Precise Spiking Patterns through Reward Modulated
  Spike-Timing-Dependent Plasticity.
\newblock In: {Artificial Neural Networks--ICANN 2013}. Springer; 2013. p.
  256--263.

\bibitem{Gerstner2002}
Gerstner W, Kistler WM.
\newblock Spiking neuron models: Single neurons, populations, plasticity.
\newblock Cambridge University Press; 2002.

\bibitem{Jolivet2006}
Jolivet R, Rauch A, L{\"u}scher HR, Gerstner W.
\newblock Predicting spike timing of neocortical pyramidal neurons by simple
  threshold models.
\newblock Journal of Computational Neuroscience. 2006;21(1):35--49.

\bibitem{Paninski2004}
Paninski L.
\newblock Maximum likelihood estimation of cascade point-process neural
  encoding models.
\newblock Network: Computation in Neural Systems. 2004;15(4):243--262.

\bibitem{Xu2013}
Xu Y, Zeng X, Han L, Yang J.
\newblock A supervised multi-spike learning algorithm based on gradient descent
  for spiking neural networks.
\newblock Neural Networks. 2013;43:99--113.

\bibitem{Rossum2001}
van Rossum MC.
\newblock A novel spike distance.
\newblock Neural Computation. 2001;13(4):751--763.

\bibitem{Gutig2006}
G{\"u}tig R, Sompolinsky H.
\newblock The tempotron: a neuron that learns spike timing--based decisions.
\newblock Nature Neuroscience. 2006;9(3):420--428.

\bibitem{Ponulak2010}
Ponulak F, Kasinski A.
\newblock Supervised learning in spiking neural networks with resume: Sequence
  learning, classification, and spike shifting.
\newblock Neural Computation. 2010;22(2):467--510.

\bibitem{Albers2013}
Albers C, Westkott M, Pawelzik K.
\newblock Perfect Associative Learning with Spike-Timing-Dependent Plasticity.
\newblock In: Advances in Neural Information Processing Systems; 2013. p.
  1709--1717.

\bibitem{Sporea2013}
Sporea I, Gr{\"u}ning A.
\newblock Supervised learning in multilayer spiking neural networks.
\newblock Neural Computation. 2013;25(2):473--509.

\bibitem{Albers2016}
Albers C, Westkott M, Pawelzik K.
\newblock Learning of Precise Spike Times with Homeostatic Membrane Potential
  Dependent Synaptic Plasticity.
\newblock PLoS ONE. 2016;11(2):e0148948.

\bibitem{Bohte2002}
Bohte SM, Kok JN, La~Poutre H.
\newblock Error-backpropagation in temporally encoded networks of spiking
  neurons.
\newblock Neurocomputing. 2002;48(1):17--37.

\bibitem{Bush2012}
Bush D, Jin Y.
\newblock Calcium control of triphasic hippocampal {STDP}.
\newblock Journal of Computational Neuroscience. 2012;33(3):495--514.

\bibitem{Knudsen1994}
Knudsen EI.
\newblock Supervised learning in the brain.
\newblock Journal of Neuroscience. 1994;14(7):3985--3997.

\bibitem{Miall1996}
Miall RC, Wolpert DM.
\newblock Forward models for physiological motor control.
\newblock Neural Networks. 1996;9(8):1265--1279.

\bibitem{Urbanczik2014}
Urbanczik R, Senn W.
\newblock Learning by the dendritic prediction of somatic spiking.
\newblock Neuron. 2014;81(3):521--528.

\bibitem{Izhikevich2007}
Izhikevich EM.
\newblock Solving the distal reward problem through linkage of {STDP} and
  dopamine signaling.
\newblock Cerebral Cortex. 2007;17(10):2443--2452.

\bibitem{Farries2007}
Farries MA, Fairhall AL.
\newblock Reinforcement Learning With Modulated Spike Timing--Dependent
  Synaptic Plasticity.
\newblock Journal of Neurophysiology. 2007;98(6):3648--3665.

\bibitem{Gruning2007}
Gr{\"u}ning A.
\newblock Elman backpropagation as reinforcement for simple recurrent networks.
\newblock Neural Computation. 2007;19(11):3108--3131.

\bibitem{Furber2014}
Furber SB, Galluppi F, Temple S, Plana LA.
\newblock The {SpiNNaker} project.
\newblock Proceedings of the IEEE. 2014;102(5):652--665.

\end{thebibliography}
\end{document}